\documentclass[conference]{IEEEtran}
\usepackage[linesnumbered,ruled,vlined]{algorithm2e}
\usepackage{cite}
\usepackage{cancel}

\usepackage{amsmath,amsfonts,amssymb,amsthm}
\usepackage{algorithmic}
\usepackage{balance}
\usepackage{booktabs}
\usepackage[font=small,labelfont=bf]{caption}
\usepackage{collcell}
\usepackage{dblfloatfix}
\let\labelindent\relax
\usepackage{enumitem}
\usepackage{etoolbox}
\usepackage{graphicx}
\usepackage{pgf}
\usepackage{tcolorbox}
\usepackage{subfig}  
\usepackage[capitalize,noabbrev,nameinlink]{cleveref} 
\crefname{lem}{Lemma}{Lemmas}
\crefname{defin}{Definition}{Definitions}
\crefname{thm}{Theorem}{Theorems}

\usepackage{makecell}
\usepackage{microtype}
\usepackage[numbers, sort&compress]{natbib}
\usepackage{pifont}
\usepackage{relsize}
\usepackage{siunitx} 

\usepackage{tikz}
\usepackage{pgfplots}
\pgfplotsset{compat=1.13}
\usepgfplotslibrary{fillbetween}
\usetikzlibrary{pgfplots.groupplots}

\usepackage{url}
\DeclareSIUnit\speedup{\mathrm{\times}}
\usepackage{textcomp}
\usepackage{xspace}
\def\BibTeX{{\rm B\kern-.05em{\sc i\kern-.025em b}\kern-.08em
		T\kern-.1667em\lower.7ex\hbox{E}\kern-.125emX}}

\definecolor{cbone}    {HTML}{006BA4} 
\definecolor{cbtwo}    {HTML}{FF800E} 
\definecolor{cbthree}  {HTML}{ABABAB} 
\definecolor{cbfour}   {HTML}{595959} 
\definecolor{cbfive}   {HTML}{5F9ED1} 
\definecolor{cbsix}    {HTML}{C85200} 
\definecolor{cbseven}  {HTML}{898989} 
\definecolor{cbeight}  {HTML}{A2C8EC} 
\definecolor{cbnine}   {HTML}{FFBC79} 
\definecolor{cbten}    {HTML}{CFCFCF} 

\definecolor{pyone}{HTML}   {1f77b4} 
\definecolor{pytwo}{HTML}   {ff7f0e} 
\definecolor{pythree}{HTML} {2ca02c} 
\definecolor{pyfour}{HTML}  {d62728} %
\definecolor{pyfive}{HTML}  {9467bd} %
\definecolor{pysix}{HTML}   {8c564b} %
\definecolor{pyseven}{HTML} {e377c2} %
\definecolor{pyeight}{HTML} {7f7f7f} %
\definecolor{pynine}{HTML}  {bcbd22} %

\newcommand{\tab}[1]{\cref{#1}}
\newcommand{\fig}[1]{\cref{#1}}
\newcommand{\sect}[1]{\cref{#1}}
\newcommand{\appdx}[1]{\cref{#1}}

\newcommand{\tablesize}{\small}

\DeclareMathOperator*{\argmin}{argmin\,}

\newcommand{\unlearn}[2]{{#1 \rightarrow #2}}
\newcommand{\point}{z}
\newcommand{\pert}{\tilde{z}}
\newcommand{\points}{Z}
\newcommand{\perts}{\tilde{Z}}
\newcommand{\loss}{\ell}
\newcommand{\gradlossPoint}{\nabla_{\model}\loss(\point, \optmodel)}
\newcommand{\gradlossPert}{\nabla_{\model}\loss(\pert, \optmodel)}
\newcommand{\bigloss}{L}
\newcommand{\optmodel}{\theta^*}
\newcommand{\model}{\theta}
\newcommand{\Model}{\Theta}
\newcommand{\feat}{f}
\newcommand{\feats}{F}
\newcommand{\val}{v}
\newcommand{\vals}{V}
\newcommand{\lipschitz}{Lipschitz continuity\xspace}

\newcommand{\txt}[1]{{\textcolor{pyone!80!black}{``#1''}}\xspace}

\renewcommand{\paragraph}[1]{\vspace{2pt}\noindent\textbf{#1.}}
\newcommand{\quasipara}[1]{\vspace{1pt}\noindent\emph{#1.}}

\newcommand{\drebin}{Malware\xspace}
\newcommand{\enron}{Spam\xspace}
\newcommand{\diabetis}{Diabetis\xspace}
\newcommand{\adult}{Adult\xspace}

\newcommand{\perc}[1]{%
	\ifstrempty{#1}%
	{\,\si{\percent}}%
	{\SI{#1}{\percent}}%
}

\SetKwInput{KwInput}{Input}           
\SetKwInput{KwOutput}{Output}         

\def\BibTeX{{\rm B\kern-.05em{\sc i\kern-.025em b}\kern-.08em
		T\kern-.1667em\lower.7ex\hbox{E}\kern-.125emX}}

\begin{document}
%
\title{Machine Unlearning of Features and Labels\\[-4mm]}



%

\author{}
\author{\IEEEauthorblockN{Alexander Warnecke\IEEEauthorrefmark{1},
Lukas Pirch\IEEEauthorrefmark{1}, 
Christian Wressnegger\IEEEauthorrefmark{2} and
Konrad Rieck\IEEEauthorrefmark{1}}
\IEEEauthorblockA{\IEEEauthorrefmark{1}Technische Universität Braunschweig}
\IEEEauthorblockA{\IEEEauthorrefmark{2} KASTEL Security Research Labs, Karlsruhe Institute of Technology (KIT)}}


\IEEEoverridecommandlockouts
\makeatletter\def\@IEEEpubidpullup{6.5\baselineskip}\makeatother
\IEEEpubid{\parbox{\columnwidth}{
    Network and Distributed System Security (NDSS) Symposium 2023\\
    28 February - 4 March 2023, San Diego, CA, USA\\
    ISBN 1-891562-83-5\\
    https://dx.doi.org/10.14722/ndss.2023.23xxx\\
    www.ndss-symposium.org
}
\hspace{\columnsep}\makebox[\columnwidth]{}}

\maketitle

\begin{abstract}
  Removing information from a machine learning model is a non-trivial
  task that requires to partially revert the training process. This
  task is unavoidable when sensitive data, such as credit card numbers
  or passwords, accidentally enter the model and need to be removed
  afterwards. Recently, different concepts for machine unlearning have
  been proposed to address this problem. While these approaches are
  effective in removing individual data points, they do not scale to
  scenarios where larger groups of features and labels need to be
  reverted.
  In this paper, we propose the first method for unlearning features
  and labels. Our approach builds on the concept of influence
  functions and realizes unlearning through closed-form updates of
  model parameters. It enables to adapt the influence of training data
  on a learning model retrospectively, thereby correcting data leaks
  and privacy issues. For learning models with strongly convex loss
  functions, our method provides certified unlearning with theoretical
  guarantees. For models with non-convex losses, we empirically show
  that the unlearning of features and labels is effective and
  significantly faster than other strategies.
\end{abstract}

\section{Introduction}
Machine learning has become an ubiquitous tool in analyzing personal
data and developing data-driven services.  Unfortunately, the
underlying learning models can pose a privacy threat if they
inadvertently capture sensitive information from the training data and
later reveal it to users. For example, \citet{CarLiuErl+19} show that
the Google text completion system contains credit card numbers from
personal emails, which may be exposed to other users during
auto-completion. In addition, privacy regulations, such as the
European GDPR~\citep{gdpr}, enable users to request the removal of
their personal data from learning models as part of the ``right to be
forgotten''.

Deleting data from a learning model is a challenging task that
requires selectively reverting the learning process.  In the absence
of specific methods, the only option is to retrain the model from
scratch, which is costly and only possible if the original data is
still available. As a remedy, \citet{CaoYan15} and
\mbox{\citet{BourChaCho+21}} propose methods for \emph{machine
  unlearning}. These methods partially reverse the learning process
and are capable of deleting learned data points in retrospection.  As
a result, they enable to mitigate privacy leaks and comply with
removal requests from users.

Information leaks, however, do not only manifest in isolated data
points. When training on content of social media, sensitive data is
often distributed across several data instances. For example, the
leaked home address of a celebrity may be shared in thousands of
posts, rendering the removal of affected data points
inefficient. Similarly, when applying machine learning on emails,
personal data in conversations, such as names, addresses, and phone
numbers, can affect dozens of messages and form features in the
learning model whose later removal requires substantial changes to the
model's structure.


Existing approaches for unlearning \citep{BourChaCho+21, CaoYan15,
  GuoGolHan+20, GinMelGua+19, NeeRotSha20, AldMahBei20} are
inefficient in these cases, as they operate on data points only:
First, a runtime improvement can hardly be obtained over retraining
when the changes are not isolated and larger parts of the data need to
be corrected. Second, removing multiple data points reduces the
fidelity of the corrected model and thus is not a viable option in
practical scenarios. Consequently, unlearning should not be limited to
removing data points, but allow corrections at different granularity
of the training data, such as fixing leaks in features and labels
individually.

In this paper, we propose the first method for unlearning features and
labels from a learning model. Our approach is inspired by the concept
of \emph{influence functions}, a technique from robust
statistics~\citep[][]{Ham74}, that allows for estimating the influence
of data on learning models~\citep[][]{KohLia17, KohSiaTeo+19}. By
reformulating this influence estimation as a form of unlearning, we
derive a versatile approach that maps changes of the training data in
retrospection to closed-form updates of model parameters. These
updates can be calculated efficiently, even if larger parts of the
training data are affected, and enable correcting features and labels
captured within the model. As a result, our method can remove privacy
leaks and other unwanted content from a wide range of common learning
models.

For models with strongly convex loss, such as logistic regression and
support vector machines, we prove that our approach enables
\emph{certified unlearning}. That is, it provides theoretical
guarantees on the removal of features and labels from the models. To
obtain these guarantees, we build on the concepts of certified data
removal~\mbox{\citep{GuoGolHan+20, NeeRotSha20}} and differential
privacy~\citep{ChaMonSar11,Dwo06}. In particular, we measure the
difference between models obtained using our approach and retraining
on corrected data. By carefully introducing noise into the learning
process, we derive an upper bound on this difference and thus realize
provable unlearning in practice.

For models with non-convex loss functions, such as deep neural
networks, similar guarantees cannot be realized. However, we
empirically demonstrate that our approach is significantly faster in
comparison to sharding~\citep{GinMelGua+19, BourChaCho+21} and
retraining while reaching a similar level of accuracy. Moreover,
due to the compact updates, our approach requires only a fraction of the
training data and hence is applicable when the original data is not
available. We demonstrate the efficacy of our approach in case studies
on unlearning (a) sensitive features in linear models, (b) unintended
memorization in language models, and (c) label poisoning in computer
vision.


\paragraph{Contributions}
In summary, we make the following major contributions in this paper:

\begin{enumerate}
	\setlength{\itemsep}{5pt}
	
	\item \emph{Unlearning with closed-form updates.} We introduce a
	novel framework for unlearning features and labels.  This
	framework builds on closed-form updates of model parameters
	and thus is significantly faster than instance-based approaches
	to unlearning.
	
	\item \emph{Certified unlearning.} We derive two unlearning
	strategies for our framework based on first-order and
	second-order gradient updates. Under convexity and continuity
	assumptions on the loss, we show that both strategies provide
	certified unlearning of data.
	
	\item \emph{Empirical analysis.}  We empirically show that
	unlearning of sensible information is possible even for deep
	neural networks with non-convex loss functions. We find that
	our first-order update is highly efficient, enabling a
	speed-up over retraining by several orders of magnitude.
	
\end{enumerate}

\paragraph{Roadmap} We review related work on machine unlearning and
influence functions in \cref{sec:relatedwork}. We motivate our
approach in \cref{sec:motivation} and introduce its technical
realizations in \cref{sec:approach} and
\ref{sec:updat-learn-models}. A theoretical analysis is presented in
\cref{sec:cert-unlearning} and an empirical evaluation in
\cref{sec:empirical-analysis}. We discuss limitations in
\cref{sec:limitations} and conclude in \sect{sec:conclusions}.

\section{Related work}

\label{sec:relatedwork}

The increasing application of machine learning to personal data has
started a series of research on detecting and correcting privacy
issues in learning models~\citep[e.g.,][]{ZanWutTop+20, CarTraWal+21,
	CarLiuErl+19, SalZhaHumBer+19, LeiFre20, ShoStrSonShm+17}.  In the
following, we provide an overview of work on machine unlearning and
influence functions. A broader discussion of privacy and machine
learning is given by~\citet{DeC21} and~\citet{PapMcDSinWel18}.

\paragraph{Machine unlearning}
Methods for removing sensitive data from learning models are a recent
branch of security research. As one of the first, \citet{CaoYan15}
show that a large number of learning models can be represented in a
closed summation form that allows for elegantly removing individual
data points in retrospection. However, for adaptive learning
strategies, such as stochastic gradient descent, this approach
provides only little advantage over retraining and thus is not well
suited for correcting problems in deep neural networks.

\citet{BourChaCho+21} address this problem and propose a strategy for
unlearning data instances from general classification
models. Similarly, \citet{GinMelGua+19} develop a technique for
unlearning points from clusterings.  The key idea of both approaches
is to split the data into independent partitions---so called
shards---and aggregate the final model from sub-models trained over
these shards. Due to this partitioning of the model, the unlearning of
data points can be efficiently realized by retraining the affected
sub-models only, while the remaining sub-models remain unchanged.
\mbox{\citet{AldMahBei20}} show that this approach can be further sped
up for least-squares regression by choosing the shards cleverly. We
refer to this family of unlearning methods as \emph{sharding}.

\begin{figure}[t!]
	\centering
	\subfloat{
        \scalebox{0.92}{\definecolor{S5}{HTML}{bf40bf}
\definecolor{S10}{HTML}{1f77b4}
\definecolor{S20}{HTML}{ff7f0e}
\definecolor{S30}{HTML}{2ca02c}

\begin{tikzpicture}
    \pgfplotsset{footnotesize,samples=10}
    \begin{groupplot}[group style = {group size = 1 by 1}, width = 0.92\linewidth, height = 0.52\linewidth]
        \nextgroupplot[
            title = {},
            ylabel={Probability},
            xlabel={Number of unlearning samples},
            ymin=0.0, ymax=1.0,
            ytick={0.0, 0.5, 1.0},
            xmin=0.0, xmax=300,
            ymode=linear,
            yminorticks=false,
            xtick={0, 100, 200, 300},
            legend to name={grouplegend}
            ]
            \addplot[color=S5, mark=none, line width=1.2pt, densely dotted] plot [] table [x=n, y=proba, col sep=comma]{figures/Unlearning/sharding/sharding_n-S5.csv};
            \addlegendentry{S = 5}%

            \addplot[color=S10, mark=none, line width=1.2pt, dash dot] plot [] table [x=n, y=proba, col sep=comma]{figures/Unlearning/sharding/sharding_n-S10.csv};
            \addlegendentry{S = 10}%

            \addplot[color=S20, mark=none, line width=1.2pt, dashed] plot [] table [x=n, y=proba, col sep=comma]{figures/Unlearning/sharding/sharding_n-S20.csv};
            \addlegendentry{S = 20}%

            \addplot[color=S30, mark=none, line width=1.2pt] plot [] table [x=n, y=proba, col sep=comma]{figures/Unlearning/sharding/sharding_n-S30.csv};
            \addlegendentry{S = 30}%
    \end{groupplot}
    \node at ($(group c1r1)+ (2.4cm ,-0.675cm)$) {\ref{grouplegend}}; 
\end{tikzpicture}}}
	\caption{Probability of all shards being affected when unlearning
		for varying number of data points and shards ($S$).}
	\label{fig:sharding}
        \vspace{-10pt} 
\end{figure}
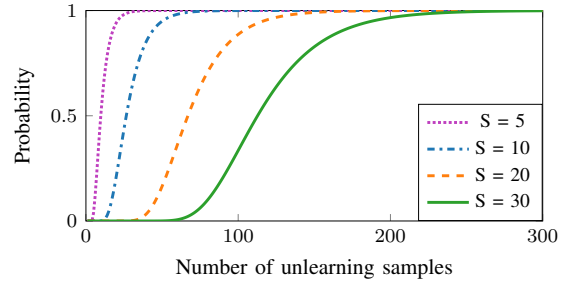

Unfortunately, sharding has one critical drawback: Its efficiency
quickly deteriorates when multiple data points need to be corrected.
The probability that all shards need to be retrained increases with
the number of affected data points, as shown in \cref{fig:sharding}.
For a practical setup with \num{20}~shards, as proposed by
\mbox{\citet{BourChaCho+21}}, changes to as few as \num{150}~points
are sufficient to impact all shards and render the approach
inefficient.  We provide a detailed analysis of this limitation
in~\sect{appdx:sharding}.

\paragraph{Influence functions}
We base our approach on influence functions, a classic concept
originating from robust statistics~\citep{Ham74}. An \emph{influence
  function} is a measure of the dependence of a statistical estimator
on the value of a data point. This concept was originally introduced
by~\citet{CooWei82} for investigating the changes of linear regression
models. Although influence functions have been occasionally employed
in machine learning~\citep[e.g.,][]{LecDenkSol90, HasStoWol94}, it was
the seminal work of \citet{KohLia17} that brought general attention to
this concept and its application to learning models. In particular,
\citeauthor{KohLia17} use influence functions for measuring the impact
of training points on the predictions of a learning model.

Influence functions have then been used in a wide range of
applications. For example, they have been applied to trace bias in
word embeddings to documents~\citep{BruAlkAnd+19, CheSiLi+20},
determine reliable regions in learning models \citep{SchuSar19}, and
explain deep neural networks~\citep{BasPopFei20}. As part of this
research strain, \citet{BasYouFei} increase the accuracy of influence
functions by using high-order approximations, \citet{BarBruDzi20}
improve their precision, and \mbox{\citet{GuoRajHas+20}} reduce their
runtime by focusing on specific samples. Finally, \citet{GolAchRav+21,
  GolAchSoa20} move to the field of privacy and use influence
functions to ``forget'' data points in neural networks using special
approximations.

In terms of theoretical analysis, \citet{KohSiaTeo+19} study the
accuracy of influence functions when estimating the loss on test data
and \citet{NeeRotSha20} perform a similar analysis for gradient-based
update strategies. In addition, \citet{RadMal} show that the
prediction error on leave-one-out validations can be reduced with
influence functions. Finally, \mbox{\citet{GuoGolHan+20}} build on the
concept of differential privacy and introduce the idea of certified
removal of single data points. They propose a definition of
indistinguishability between learning models similar
to~\citet{NeeRotSha20}.  In this paper, we expand this work to
certified unlearning of features and labels.

\section{Problem Setting and Threat Model}
\label{sec:motivation}

Before presenting the technical realization of our approach to
unlearning, let us first describe the problem setting along with the
underlying threat model.

\paragraph{Threat model}
For our approach, we consider learning models trained on
privacy-sensitive data and accessible to users through an interface,
such as a text completion system, a learning chatbot, or a
collaborative spam filter. As these models operate on sensitive data,
there is a need to protect their users' privacy and to close leaks in
their interfaces as soon as possible.

\begin{enumerate}
\setlength{\itemsep}{4pt}

\item First, this need arises from privacy regulations, such as the
  European GDPR. According to the GDPR, European citizens can request
  their personal information to be removed from a service to protect their
  privacy, including derived data in learning
  models~\citep{gdpr}.

\item Second, unintended memorization of personal data, such as credit
  card numbers, may also demand mechanisms for its removal. For
  example, recent work shows that text completion systems allow
  adversaries to extract sensitive data through their
  interfaces~\citep{CarLiuErl+19,CarTraWal+21}.

\item Third, data used for constructing a learning model may
  accidentally violate ethical standards and thus also require removal
  once these violations have been detected. For example, the chatbot
  ``Tay'' by Microsoft memorized and replicated anti-semitic and
  racist content~\citep{web:Tay}.

\end{enumerate}

In all these cases, service providers must mitigate the resulting
threats and immediately correct the learning model to reduce potential
harm to their users.

\paragraph{From retraining to unlearning}
At a first glance, retraining from scratch may seem like the optimal
strategy to fix issues in learning models. By correcting the training
data directly, all of the above issues can be reliably
resolved. However, retraining from scratch comes with disadvantages:

\begin{enumerate}
\setlength{\itemsep}{4pt}

\item Depending on the size of the original data, retraining from
  scratch can be costly. The entire learning process needs to be
  reproduced even if only a few data instances, features, or labels
  need to be corrected.

\item Privacy regulations require that the purpose and duration of
  data storage are clearly defined and approved by the
  user. Therefore, it may not be possible to keep the original data
  indefinitely for retraining.

\item Finally, with an online learning system like a chatbot, training
  data is volatile and might not be fully available. Nevertheless,
  even with such systems, inappropriate memorization must be removed
  quickly.

\end{enumerate}

As a result of this situation, different concepts for machine
unlearning have been proposed in the last years \citep{BourChaCho+21,
  CaoYan15, GuoGolHan+20, GinMelGua+19, NeeRotSha20, AldMahBei20}.  We
follow this line of work and introduce a new mechanism for removing
features and labels from learning models.

\paragraph{Unlearning instances vs. features}
In many learning-based systems, data points are directly linked to
individuals. For example, in a face recognition system, the training
data consists of portrait photos, each showing one person. In this
scenario, unlearning is naturally performed on the level of data
points. However, privacy issues can also arise at a different
granularity of the data. For example, the leaked address of a
celebrity may be widely circulated on social media, affecting features
of hundreds of data points. The same problem occurs when training on
emails and sensitive data, such as personal names or telephone
numbers, appears in several emails during a conversation. Similarly,
toxic content may be captured by language models from posts of various
users over time.

\begin{figure}[t]
	\centering
        \includegraphics[width=0.44\textwidth]{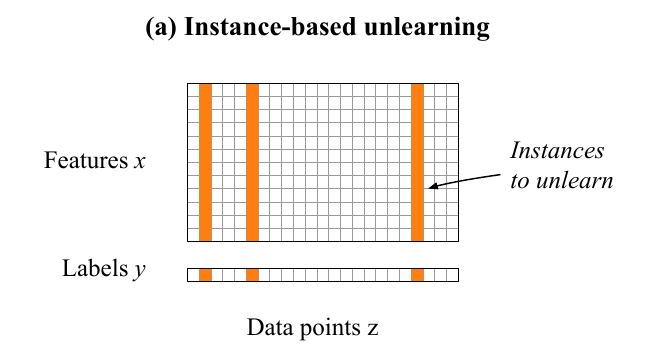}\\[5pt]
        \includegraphics[width=0.44\textwidth]{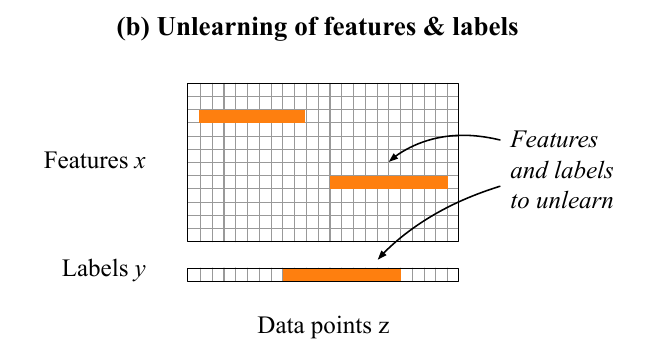}
	\caption{Instance-based unlearning vs. unlearning of features and
		labels.  The data to be removed is marked with orange. }
	\label{fig:unlearning-diff}
        \vspace{-8pt}
\end{figure}

In these cases, instance-based unlearning \citep{BourChaCho+21,
  CaoYan15, GuoGolHan+20, GinMelGua+19, NeeRotSha20, AldMahBei20} is
inefficient: First, the runtime advantage over retraining vanishes
with the number of data points affected, as shown in the previous
section. Second, removing entire instances and not just the affected
features or labels unnecessarily degrades the performance of the
corrected models. As a solution to this situation, we propose a method
for unlearning of individual features and labels. As illustrated in
Figure~\ref{fig:unlearning-diff}, this strategy operates on an
orthogonal dimension of the data. Instead of correcting privacy issues
along data instances (columns), we focus on resolving the issues in
feature values and labels (rows).
This is possible because we formulate unlearning as a closed-form
update of the model, which enables us to correct features and labels
at arbitrary positions in the training data.
%


To the best of our knowledge, we are the first to tackle the problem
of unlearning from this perspective, thereby adding a new tool to the
existing machinery for mitigating privacy threats in machine
learning. In particular, our approach provides the following advantages:

\begin{itemize}
\setlength{\itemsep}{5pt}

\item \emph{Efficiency.} When privacy issues affect multiple data
  instances but are limited to particular features or labels, it is
  more efficient to correct these directly. As demonstrated in our
  evaluation, we achieve a significant performance improvement over
  existing approaches.

\item \emph{Flexibility.} Due to the concept of influence functions,
  we can correct arbitrary feature values and labels in the original
  training data. As a result, our approach can also unlearn entire
  data points, making it a versatile alternative to existing methods.

\end{itemize}


\section{Unlearning with Influence}
\label{sec:approach}

As basis for our approach, we consider a supervised learning task that
is described by a dataset $D=\{\point_1,\dots,\point_n\}$ with each
point $\point_i=(x, y)$ consisting of features $x \in \mathcal{X}$ and
a label $y \in \mathcal{Y}$. We assume that
$\mathcal{X} = \mathbb{R}^d$ and denote the $j$-th feature of $x$ by
$x[j]$.  Given a loss function $\loss(\point,\model)$ that measures
the difference between the predictions of a learning model~$\model$
and the true labels, the optimal model~$\optmodel$ can be found by
minimizing the regularized empirical~risk,
\begin{align}
	\optmodel &=
	\argmin_{\model}\bigloss_b(\model; D) \nonumber \\ &=
	\argmin_{\model} \sum_{i=1}^n \loss(\point_i, \model)+
	\lambda \Omega(\model)
	+ b^T\theta
	\label{eq:emrisk}
\end{align}
where $\bigloss_b$ is the loss on the entire dataset $D$ and $\Omega$
a common regularizer \citep[see][]{DudHarSto00}. Note that we add a
vector \mbox{$b\in\mathbb{R}^p$} to the optimization. For conventional
learning, this vector is set to zero and can be ignored. For realizing
certified unlearning, however, it enables to add a small amount of
noise to the minimization, similar to differentially private
learning~\mbox{\citep{Dwo06,DwoRot14}}.  We introduce this technique
later in Section~\ref{sec:cert-unlearning} and omit the subscript in
$\bigloss_b$ for now.


\subsection{Unlearning Data Points}

We begin the design of our approach by asking a simple question: How
would the optimal model~$\optmodel$ change, if only one data
point~$\point$ had been perturbed by some change~$\delta$?  Replacing
$\point$ by $\pert = (x+\delta, y)$ leads to the new optimal model:
\begin{equation}
	\optmodel_{\unlearn{\point}{\pert}} =
	\argmin_{\model}
	\bigloss(\model; D) + \loss(\pert, \model) - \loss(\point, \model).
	\label{eq:eq1}
\end{equation}
However, calculating the new model
$\optmodel_{\unlearn{\point}{\pert}}$ exactly is expensive and does
not provide any advantage over solving the problem in
\cref{eq:emrisk}.  Instead of replacing the data point~$\point$ with
$\pert$, we can also up-weight $\pert$ by a small value~$\epsilon$ and
down-weight $\point$ accordingly, resulting in the following problem:
\begin{align}
	\optmodel_{\epsilon, \unlearn{\point}{\pert}} =
	\argmin_{\model}& \, \bigloss(\model; D)
	+ \epsilon \loss(\pert, \model)
	- \epsilon \loss(\point, \model). \label{eq:eq2}
\end{align}
\cref{eq:eq1,eq:eq2} are equivalent for $\epsilon=1$ and solve the
same problem.

As a result, we do not need to explicitly remove a data point from the
training data but can revert its \emph{influence} on the learning
model through a combination of up-weighting and down-weighting.
It is easy to see that this approach is not restricted to a single
point. We can define a set of points~$\points$ as well as their
perturbed versions~$\perts$. 
That is, $\points$ contains the original data, while $\perts$ its
corrected version.
Based on this definition, w
arrive at the following optimization problem:
\begin{align}
	\optmodel_{\epsilon, \unlearn{\points}{\perts}}
	= \argmin_{\model} & \bigloss(\model; D) \, + \nonumber\\
	& \epsilon \sum_{\pert \in \perts} \loss(\pert, \model)
	- \epsilon \sum_{\point \in \points} \loss(\point, \model).
	\label{eq:eq2a}
\end{align}
This generalization enables us to approximate changes on larger parts
of the training data. Instead of solving the problem in
\cref{eq:eq2a}, however, we formulate the optimization as an update of
the original model $\optmodel$. That is, we seek a closed-form update
$\Delta(\points, \perts)$ of the model parameters, such that
\begin{equation}
	\label{eqn:update_delta}
	\optmodel_{\epsilon, \unlearn{\points}{\perts}} \approx 
	\optmodel + \Delta(\points, \perts),
\end{equation}
where $\Delta(\points, \perts)$ has the same dimension as the learning
model~$\model$ but is sparse and affects only the necessary weights.

As a result of this formulation, we can describe changes of the
training data as a compact update $\Delta$.
 We show in \sect{sec:updat-learn-models} that
this update step can be efficiently computed using first-order and
second-order derivatives. 
Note that if $\perts = \emptyset$ in Equation~\eqref{eq:eq2a}, our
approach also yields updates to remove \emph{data points} similar to
prior work~\citep{KohLia17, GuoGolHan+20}.

\subsection{Unlearning Features and Labels}
\label{sec:approach_2}

Equipped with a general method for updating a learning model, we
proceed to introduce our approach for unlearning features and
labels. To this end, we expand our notion of perturbations and include
changes to labels by defining
\begin{equation*}
	\pert = (x + \delta_x, y + \delta_y),
\end{equation*}
where $\delta_x$ modifies the features of a data point and $\delta_y$
its label. By specifying different perturbations $\perts$, we can now
realize several unlearning tasks with closed-form updates.

\paragraph{Replacing features}
As the first type of unlearning, we consider the task of correcting
features in a learning model. This task is relevant if the content of
some features violates the privacy of a user and needs to be replaced
with alternative data. As an example, personal names, home addresses,
or other sensitive information might need to be removed after a model
has been trained on a corpus of emails. Similarly, in a credit scoring
system, the race, gender or other biasing features might need to be
replaced with neutral content.


For a set of features $\feats$ and their new values $\vals$, we define
perturbations on the affected points $\points$ by
\begin{equation*}
	\perts=\big \{ (x[\feat]=\val, y) :
	(x,y) \in \points, \,
	(\feat,\val)\in \feats \times \vals \big \}.
\end{equation*}
%
For example, a credit card number contained in the training data can
be blinded by a random number sequence in this setting. The values
$\vals$ can be adapted individually for each data point, so that
fine-grained corrections are possible.

\paragraph{Replacing labels} As the second type of unlearning, we
focus on correcting labels. This form of unlearning is necessary if
the labels captured in a model contain unwanted or inappropriate
information.  For example, in generative language models, the training
text is used as input features (preceding tokens) \emph{and}
labels (target~tokens) \citep{Gra13,SutMarHin11}. Hence, defects
can only be eliminated if the labels are unlearned as well.
%

For the affected points $\points$ and the set of new labels~$Y$, we
define the corresponding perturbations by
\begin{equation*}
	\perts=\big \{ (x, y) \in \points_x \times Y \big \},
\end{equation*}
where $\points_x$ corresponds to the data points in $\points$ without
their original labels. The new labels $Y$ can also be individually
selected for each data point, as long as they come from the
domain~$\mathcal{Y}$, that is, $Y \subset \mathcal{Y}$.
Note that the replaced labels and features can be easily combined in
one set of perturbations $\perts$, so that defects affecting both can
be corrected in a single update. In \sect{sec:unle-unint-memor}, we
demonstrate that this combination can be used to remove unintended
memorization from generative language models with high efficiency.

\paragraph{Revoking features} Based on appropriate definitions of
$\points$ and~$\perts$, our approach enables to replace the content of
features and thus eliminate privacy leaks by overwriting sensitive
data. In some scenarios, however, it might be necessary to even
completely remove features from a learning model---a task that we
denote as \emph{revocation}. In contrast to the correction of
features, this form of unlearning poses a unique challenge: The
revocation of features reduces the input dimension of the model.
While this adjustment can be easily carried out through retraining
with adapted data, constructing a model update as in
\cref{eqn:update_delta} becomes tricky.

To address this problem, let us consider a model~$\optmodel$ trained
on a dataset $D\subset\mathbb{R}^d$. If we remove some features
$\feats$ from this dataset and train the model again, we obtain a new
optimal model $\optmodel_{-\feats}$ with reduced input dimension. By
contrast, if we set the values of the features $\feats$ to zero in the
dataset and train again, we obtain an optimal model
$\optmodel_{\feats=0}$ with the same input dimension as~$\optmodel$.
These two models are equivalent for a large class of learning models,
including several neural networks as the following lemma shows.

\newtheorem{lem}{Lemma}
\begin{lem}
	\label{lemma1}
	For learning models processing inputs $x$ using linear
	transformations of the form $\model^{T} x$, we have
	$\optmodel_{-\feats} \equiv \optmodel_{\feats=0}$.
\end{lem}

\begin{proof}
	It is easy to see that it is irrelevant for the dot product
	$\model^Tx$ whether a dimension of $x$ is missing or equals zero in
	the linear transformation
	\begin{equation*}
		\sum_{\substack{k: k\notin \feats}}\model[k] x[k]
		= \sum_{k}\model[k] \mathbf{1}\{k\notin \feats\}x[k].
	\end{equation*}
	As a result, the loss $\loss(\point,\model)=\loss(\model^Tx,y,\model)$
	of both models is identical for every data point $\point$. Hence,
	$\bigloss(\theta;D)$ is also equal for both models and thus the
	same objective is minimized during learning, resulting in equal
	model parameters.
\end{proof}

\cref{lemma1} enables us to erase features from many learning models
by first setting them to zero, calculating the parameter update, and
then reducing the input dimension of the models
accordingly. 
Concretely, to revoke the features $\feats$ from a learning model, we
first locate all data points where these features are non-zero with
\begin{equation*}
	\points=\big \{ (x,y) \in D : x[\feat] \neq 0, \feat \in \feats
	\big \}.
\end{equation*}
Then, we construct corresponding perturbations so that the features
are set to zero with our approach,
\begin{equation*}
	\perts=\big \{ (x[\feat]=0, y) :
	(x,y) \in \points, \,
	\feat \in \feats \big \}.
\end{equation*}
Finally, we adapt the input dimension by removing the affected inputs
of the learning models, such as the corresponding neurons in the input
layer of a neural network.

%

\section{Update Steps for Unlearning}
\label{sec:updat-learn-models}

Our approach rests on changing the influence of training data in a
closed-form update. In the following, we derive two strategies for
calculating this closed form: a \emph{first-order update} and a
\emph{second-order update}. The first strategy builds on the gradient
of the loss function and thus can be applied to any model with
differentiable loss. The second strategy incorporates second-order
derivatives which limits the application to loss functions with an
invertible Hessian matrix.

\subsection{First-Order Update}
\label{sec:first-order}

Recall that we aim to find an update $\Delta(\points, \perts)$ that we
can add to our model $\optmodel$ for unlearning. If the loss $\loss$
is differentiable, we can compute an optimal \emph{first-order update}
as follows
\begin{equation}
	\label{eqn:update1}
	\Delta(\points, \perts) = -\tau \Big(\sum_{\pert \in \perts}
	\gradlossPert - \sum_{\point \in \points} \gradlossPoint\Big)
\end{equation}
where $\tau$ is a small constant that we refer to as \emph{unlearning
  rate}. A complete derivation of \cref{eqn:update1} is given in
\appdx{sec:derivation-gradient}. Intuitively, this update shifts the
model parameters from
$\sum_{\point \in \points}\nabla\loss(\point,\optmodel)$ to
$\sum_{\pert \in \perts}\nabla\loss(\pert, \optmodel)$ where the size
of the update step is determined by the rate $\tau$. This update
strategy is similar to a gradient descent update GD given by
\begin{equation*}
	\text{GD}(\perts) = -\tau \sum_{\pert \in \perts}\gradlossPert.
\end{equation*}
However, it differs from this update step in that it moves the model to the
\emph{difference} in gradient between the original and perturbed data,
which minimizes the loss on $\pert$ and at the same time removes the
information contained in $\point$.

The first-order update is a simple and yet effective strategy: The
gradients of $\loss$ can be computed in $\mathcal{O}(p)$~\citep{Pea94}
where $p$ is the number of parameters in the learning model.  
%
%
However, the first-order update involves a parameter $\tau$ that
controls the impact of the unlearning. To ensure that the data has
been completely replaced, it is necessary to calibrate this parameter
using a measure for the success of unlearning.
%
In \cref{sec:empirical-analysis}, we show how the exposure metric
proposed by \mbox{\citet{CarLiuErl+19}} can be used for this
calibration.


\subsection{Second-Order Update}
\label{sec:second-order-update}

The unlearning rate $\tau$ can be eliminated if we make further
assumptions on the properties of the loss $\loss$. If we assume that
$\loss$~is twice differentiable and strictly convex, the influence of
a single data point can be approximated in closed form by
\begin{equation*}
	\frac{\partial\optmodel_{\epsilon, \unlearn{\point}{\pert}}}%
	{\partial\epsilon}\Big \lvert_{\epsilon=0} = 
	-H_{\optmodel}^{-1}\big(\gradlossPert - \gradlossPoint\big),
\end{equation*}
where $H_{\optmodel}^{-1}$ is the inverse Hessian of the loss at
$\optmodel$, that is, the inverse matrix of the second-order partial
derivatives \citep[see][]{CooWei82}. We can then perform a linear
approximation as follows
\begin{equation}
	\optmodel_{\unlearn{\point}{\pert}} 
	\approx \optmodel - H_{\optmodel}^{-1}\big(
	\gradlossPert - \gradlossPoint\big).
	\label{eq:eq3}
\end{equation}
Since all operations are linear, we can extend \cref{eq:eq3} to
multiple data points and finally obtain the \emph{second-order update}
for our approach:
\begin{equation}
	\label{eqn:update2}
	\Delta(\points, \perts) = -H_{\optmodel}^{-1}\Big(
	\sum_{\pert \in \perts}\gradlossPert - 
	\sum_{\point \in \points}\gradlossPoint\Big).
\end{equation}
A full derivation of this update step is provided in
\appdx{sec:derivation-influence}.  Note that the update does not
require a parameter calibration, since the parameter weighting of the
changes is directly derived from the inverse Hessian of the loss
function.

The second-order update is the preferred strategy for unlearning on
models with a strongly convex and twice differentiable loss function
that guarantee the existence of $H_{\optmodel}^{-1}$.  Technically,
the update step in \cref{eqn:update2} can be easily calculated with
common machine learning frameworks. In contrast to the first-order
update, however, this computation involves the inverse Hessian matrix,
which can be difficult to construct for large learning models.


\paragraph{Calculating the inverse Hessian}
Given a model $\theta\in\mathbb{R}^p$ with $p$ parameters, forming and
inverting the Hessian requires \mbox{$\mathcal{O}(np^2+p^3)$} time and
$\mathcal{O}(p^2)$~space~\citep{KohLia17}. For models with a small
number of parameters, the matrix can be pre-computed and explicitly
stored, such that each subsequent request for unlearning only involves
a simple matrix-vector multiplication. For example, in
\cref{sec:unle-sens-names}, we demonstrate that unlearning features
from a linear model with about \num{2000}~parameters can be realized
with this approach in less than a second.

For complex learning models, such as deep neural networks, the Hessian
matrix quickly becomes too large for explicit storage. 
Still, we can approximate the inverse Hessian using a technique
proposed by \citet{KohLia17}. Its derivation and an algorithm for its
implementation are presented in~\appdx{appdx:algorithm}. While this
approximation weakens the theoretical guarantees of our approach, it
still enables successfully unlearning data from large learning models.
In \sect{sec:unle-unint-memor}, we demonstrate that this strategy can
be used to calculate second-order updates for a recurrent neural
network with \num{3.3}~million parameters in less than 30 seconds.

\section{Certified Unlearning}
\label{sec:cert-unlearning}

Machine unlearning aims at reliably removing privacy issues and
sensitive data from learning models. This task should ideally build on
theoretical guarantees to enable \emph{certified unlearning}, where
the corrected model is stochastically indistinguishable from the one
created by retraining. In the following, we derive conditions under
which the second-order updates of our approach provide certified
unlearning. To this end, we build on the concepts of
\emph{differential privacy}~\citep{Dwo06} and \emph{certified data
  removal}~\citep{GuoGolHan+20}, and adapt them to the unlearning
task.

Let $\mathcal{A}$ be a learning algorithm that outputs a model
$\model\in\Model$ after training on a dataset $D$, that is,
$\mathcal{A}:D\rightarrow\Model$. Randomness added by $\mathcal{A}$
induces a probability distribution over the output models
in~$\Model$. Moreover, we consider an unlearning method $\mathcal{U}$
that maps a model $\model$ to a corrected model
$\model_\mathcal{U} = \mathcal{U}(\model, D, D^\prime)$ where
$D^\prime$ denotes the dataset containing the perturbations $\perts$
required for the unlearning task. To measure the difference between a
model trained on $D'$ and one obtained by $\mathcal{U}$ we introduce
the concept of \emph{$\epsilon$-certified unlearning} as follows

\newtheorem{defin}{Definition}
\begin{defin}
	\label{def-ecr}
	Given some $\epsilon >0$ and a learning algorithm
	$\mathcal{A}$, an unlearning method~$\mathcal{U}$ is
	$\epsilon$-certified if
	\begin{equation*}
		e^{-\epsilon} \leq
		\frac{P\Big(\mathcal{U}\big(\mathcal{A}(D),D,D^\prime\big)
			\in\mathcal{T}\Big)}
		{P\big(\mathcal{A}(D^\prime)\in\mathcal{T}\big)}\leq e^{\epsilon}
		\label{eqn:ecr}
	\end{equation*}
	holds for all
	$\mathcal{T}\subset \Model, D, \text{and } D^\prime$.
\end{defin}

This definition ensures that the probability to obtain a model using
the unlearning method $\mathcal{U}$ and training a new model on
$D^\prime$ from scratch deviates at most by $\epsilon$. Following the
work of~\mbox{\citet{GuoGolHan+20}}, we introduce
\emph{$(\epsilon, \delta)$-certified unlearning}, a relaxed version of
$\epsilon$-certified unlearning, defined as follows.

\begin{defin}
	\label{def-edcr}
	Under the assumptions of \cref{def-ecr}, an
	unlearning method $\mathcal{U}$ is
	$(\epsilon,\delta)$-certified if
	\begin{gather*}
		P\Big(\mathcal{U}\big(\mathcal{A}(D),D,D^\prime\big)
		\in\mathcal{T}\Big)
		\leq e^{\epsilon}P\big(\mathcal{A}(D^\prime)\in\mathcal{T}\big)+
		\delta \\
		\text{and}\\
		P\big(\mathcal{A}(D^\prime)\in\mathcal{T}\big) \leq e^{\epsilon}
		P\Big(\mathcal{U}\big(\mathcal{A}(D),D,D^\prime\big)\in
		\mathcal{T}\Big)+\delta
	\end{gather*}
	hold for all
	$\mathcal{T}\subset \Model, D, \text{and } D^\prime$.
\end{defin}
This definition allows the unlearning method $\mathcal{U}$ to slightly
violate the conditions from \cref{def-ecr} by a
constant~$\delta$. Using this relaxation, it becomes possible to derive
practical conditions under which our approach realizes certified
unlearning.

\subsection{Certified Unlearning of Features and Labels}
\label{sec:cert-unle-feat} 

To construct theoretical guarantees for our approach, we make two
basic assumptions on the employed learning algorithm: First, we assume
that the loss function~$\loss$ is twice differentiable and strictly
convex, so that $H^{-1}$ always exists and the second-order update is
applicable.  Second, we consider an \mbox{$L_2$ regularization} in the
optimization problem \eqref{eq:emrisk}, that is, the regularizer
$\Omega(\model)$ is given by $\frac{1}{2}\Vert\model\Vert^2_2$. Both
assumptions are satisfied by a wide range of learning models,
including logistic regression and support vector machines.

A helpful tool for analyzing the task of unlearning is the
\emph{gradient residual} $\nabla\bigloss(\theta;D^\prime)$ for a given
model $\model$ and a corrected dataset~$D^\prime$. For strongly convex
loss functions, the gradient residual is zero \emph{if and only if}
$\model$ equals $\mathcal{A}(D^\prime)$ since in this case the optimum
is unique. Therefore, the norm of the gradient residual
$\Vert \nabla\bigloss(\theta;D^\prime)\Vert_2$ reflects the distance
of a model~$\model$ from the one obtained by retraining on the
corrected dataset~$D^\prime$.
%
%
%
%
We differentiate between the gradient residual of the plain loss
$\bigloss$ and the adapted loss $\bigloss_b$ where a random vector $b$
is added. The gradient residual $r$ of $\bigloss_b$ is given by
$$r=\nabla\bigloss_b(\theta ; D^\prime) =
\sum_{\point\in D^\prime}^n \nabla \loss(z,\theta)+\lambda \model +b$$
and differs from the gradient residual of $\bigloss$ only by the added
vector $b$. As a result, we can adjust the probability distribution of
$b$ to realize certified unlearning, similar to sensitivity
methods~\citep{DwoRot14}. The corresponding proofs are given in
\appdx{sec:proofs}. Moreover, for a detailed discussion on Lipschitz
continuity used in the following theorem, we refer the reader to the
paper by \citet{ChaMonSar11}.

\newtheorem{thm}{Theorem}
\begin{thm}
	\label{thm:thm1}
	Assume that $\Vert x_i\Vert_2 \leq 1$ for all data points and the
gradient
$\nabla\loss(z,\model)$ is $\gamma_z$-Lipschitz with respect to $z$ at
$\optmodel$ and \mbox{$\gamma$-Lipschitz} with respect to
$\model$. Further let $\perts$ change the features $j,\dots,j+F$ by
magnitudes at most $m_j,\dots,m_{j+F}$. If $M=\sum_{j=1}^{F} m_j$ the
following upper bounds hold:
	
\begin{enumerate}
	\item For the first-order update of our approach, we have
	    $$\big\Vert\nabla\bigloss\big(\optmodel_
		{\points\rightarrow\perts}, D^\prime\big)\big\Vert_2 \leq
		(1+\tau\gamma n)\gamma_zM\vert\points\vert$$
	\item If $\nabla^2\loss(z,\model)$ is $\gamma''$-Lipschitz with
	respect to $\model$, we have 
		$$\big\Vert\nabla\bigloss\big(\optmodel_
		{\points\rightarrow\perts}, D^\prime\big)\big\Vert_2 \leq \gamma''\Big(\frac{M\gamma_z}{\lambda}\Big)^2 n\vert\points\vert^2$$
		for the second-order update of our approach.

\end{enumerate}

\end{thm}

In order to obtain a small gradient residual norm for the first-order
update the unlearning rate should be small, ideally in the order of
$1/n\gamma$. Since $\Vert x_i\Vert_2\leq 1$ we also have
$m_j\ll1$ if $d$ is large and thus $M$ acts as an additional damping
factor for both updates when changing or revoking features.

\cref{thm:thm1} enables us to quantify the difference between
unlearning and retraining. Concretely, if $\mathcal{A}(D^\prime)$ is
an exact minimizer of $\bigloss_b$ on $D^\prime$ with density
$f_\mathcal{A}$ and $\mathcal{U}(\mathcal{A}(D), D,D^\prime)$ an
approximated minimum obtained through unlearning with density
$f_{\mathcal{U}}$, then \citet{GuoGolHan+20} show that the
max-divergence between $f_\mathcal{A}$ and $f_{\mathcal{U}}$ for the
model $\model$ produced by $\mathcal{U}$ can be bounded using the
following theorem.

\begin{thm}[\citet{GuoGolHan+20}]
	\label{thm:thm2}
	Let $\mathcal{U}$ be an unlearning method with a gradient
	residual $r$ with $\Vert r\Vert_2\leq \epsilon'$. If the
	vector $b$ is drawn from a probability distribution with
	density $p$ satisfying that for any $b_1, b_2\in\mathbb{R}^d$
	there exists an $\epsilon>0$ such that
	$\Vert b_1-b_2\Vert\leq \epsilon'$ implies
	$e^{-\epsilon}\leq \frac{p(b_1)}{p(b_2)}\leq e^{\epsilon}$
	then
	$$e^{-\epsilon}\leq\frac{f_{\mathcal{U}}(\model)}
	{f_\mathcal{A}(\model)}\leq e^{\epsilon}$$
	for any $\model$ produced by the unlearning method
	$\mathcal{U}$.
\end{thm}

\cref{thm:thm2} equips us with a way to prove the certified unlearning
property from \cref{def-ecr}. Using the gradient residual bounds
derived in \cref{thm:thm1}, we can adjust the density function
underlying the vector $b$ so that \cref{thm:thm2} holds for both
update steps of our unlearning approach.

\begin{thm}
	\label{thm:thm3}
	Let $\mathcal{A}$ be the learning algorithm that returns the
unique minimum of $\bigloss_b(\theta;D^\prime)$ and let
$\mathcal{U}$ be an unlearning method that produces a model
$\model_{\mathcal{U}}$. If
$\Vert \nabla\bigloss(\theta_\mathcal{U};D^\prime)\Vert_2 \leq
\epsilon'$
for some $\epsilon' >0$ we have the following guarantees.
\begin{enumerate}
	\item If $b$ is drawn from a distribution with density
	$p(b)=e^{-\frac{\epsilon}{\epsilon'}\Vert b\Vert_2}$ then
	$\mathcal{U}$ performs $\epsilon$-certified unlearning for
	$\mathcal{A}$.
	
	\item If $p\sim \mathcal{N}(0, c\epsilon'/\epsilon)^d$ for
	some $c>0$ then $\mathcal{U}$ performs $(\epsilon,
	\delta)$-certified unlearning for $\mathcal{A}$ with
	$\delta=1.5e^{-c^2/2}$.
\end{enumerate}
\end{thm}

\cref{thm:thm3} finally allows us to establish certified unlearning of
features and labels in practice: Given a learning model with a bounded
gradient residual norm and a privacy budget $(\epsilon, \delta)$ we
can calibrate the probability distribution of $b$ to obtain a
certified unlearning method.

\paragraph{Further details} We refer the reader for further details on
certified unlearning to the appendix. In particular, we present the
proofs for all theorems in Appendix~\ref{sec:proofs}, we discuss the
relation between our approach and differential privacy in
\cref{sect:relation-dp}, and we show how the privacy budget
$(\epsilon, \delta)$ can support multiple unlearning requests in
\cref{sect:multiple-steps}.


\section{Empirical Analysis}
\label{sec:empirical-analysis}

We proceed with an empirical analysis of our approach and its
capabilities. For this analysis, we examine the performance of
unlearning in different scenarios and compare our method to other
strategies for removing data, such as retraining, sharding,
fine-tuning, and differentially private learning. As part of these
experiments, we employ models with convex and non-convex loss
functions to understand how this property affects the success of
unlearning.


\paragraph{Unlearning scenarios} Our empirical analysis is based on
three application scenarios in which sensitive and personal
information need to be removed from learning models.


\quasipara{Scenario 1: Sensitive features} Our first scenario deals
with linear models for classification. These models are widely used in
fraud, spam and malware detection due to their simplicity and strongly
convex loss function~\citep{AttWeiDasSmo+09, ArpSprHueGasRie13}. While
they induce a slight performance drop compared to neural networks
(see~\cref{sec:feat-unle-neur}), they still remain of practical
relevance.  We investigate how our approach can unlearn sensitive
features from these models (see~\sect{sec:unle-sens-names}).

\quasipara{Scenario 2: Unintended memorization} In the second
scenario, we consider the problem of unintended memorization
\citep{CarLiuErl+19}. Language models based on recurrent neural
networks can accidentally memorize sensitive data, such as credit
card numbers or private messages.  Through specifically crafted
inputs, an attacker can extract this data during text
completion~\citep{CarTraWal+21,CarLiuErl+19}. We apply unlearning of
features and labels to remove these privacy leaks from language models
(see~\sect{sec:unle-unint-memor}).

\quasipara{Scenario 3: Data poisoning} For the third scenario, we
focus on poisoning attacks in computer vision. Here, an adversary aims
at misleading an object recognition task by flipping a few labels of
the training data. The label flips significantly reduce the
performance of the learning model. We use unlearning of labels as a
strategy to correct this defect and restore the original performance
without retraining (see~\sect{sec:unle-pois}).

\paragraph{Performance measures}
The success of unlearning depends on three properties: An effective
method must (1) remove the selected data, (2) preserve the model's
quality, and (3) be efficient compared to retraining. A method that
fails to satisfy any of these properties is ineffective, because it
either does not correctly unlearn data, degrades the model,
or lacks behind retraining. To reflect this setting, we introduce
three performance measures for our empirical analysis.

\quasipara{Efficacy of unlearning} The most important property for
successful unlearning is the removal of data. While certified
unlearning ensures this removal, we cannot provide similar guarantees
for models with non-convex loss functions. As a result, we need to
employ measures that quantitatively assess the \emph{efficacy} of
unlearning.
%
For example, we can use the \emph{exposure
  metric}~\citep{CarLiuErl+19} to measure the memorization of specific
sequences in language models after unlearning.

\quasipara{Fidelity of unlearning} The second property contributing to
the success of unlearning is the performance of the corrected model,
which we denote as \emph{fidelity}. An unlearning method is of
practical use only if it keeps the performance as close as possible to
the original model. Hence, we consider the fidelity as the second
performance measure. In our experiments, we use the loss and accuracy
of the original and corrected model on a hold-out set to determine
this property.

\quasipara{Efficiency of unlearning} If the training data used to
generate a model is available, a simple unlearning strategy is
retraining. This strategy, however, involves significant runtime and
storage costs. Therefore, we consider the \emph{efficiency} of
unlearning as the third property. In our experiments, we measure the
runtime and the number of gradient calculations for each unlearning
method on the datasets of the three scenarios.


\newcommand{\retraining}{Retraining\xspace}
\newcommand{\diffp}{Differential privacy}
\newcommand{\dpShort}{Diff. privacy}
\newcommand{\finetuning}{Fine-tuning\xspace}
\newcommand{\firstorder}{First-Order\xspace}
\newcommand{\secondorder}{Second-Order\xspace}
\newcommand{\naive}{Occlusion\xspace}
\newcommand{\sharding}[1]{SISA (#1 shards)\xspace}

\paragraph{Baseline methods} To compare our approach with prior work
on machine unlearning, we employ different baselines as reference for
examining the efficacy, fidelity, and efficiency. In particular, we
consider retraining, fine-tuning, differential privacy and sharding as
baselines.

\quasipara{\retraining} As the first baseline, we employ retraining
from scratch. This basic method is applicable if the original training
data is available and guarantees a proper removal of data. However,
the approach is costly and serves as general upper bound for the
runtime of any unlearning method.


\quasipara{\diffp\xspace (DP)} As a second baseline, we consider a
differentially private learning model \citep{ChaMonSar11}. As
discussed in \sect{sect:relation-dp}, the presence of the noise term
$b$ in our model $\optmodel$ induces a differential-privacy guarantee
which is sufficient for certified unlearning.  Therefore, we evaluate
the performance of $\optmodel$ without any following unlearning steps.

\quasipara{\finetuning} This baseline simply continues to train a
model using corrected data. We implement this fine-tuning by
performing stochastic gradient descent over the training data for one
epoch. This naive unlearning strategy serves as a middle ground
between costly retraining and specialized unlearning methods, such as
SISA and our approach.


\quasipara{SISA (Sharding)} As the fourth baseline, we consider the
unlearning method SISA proposed by \citet{BourChaCho+21}. The method
splits the training data in shards and trains sub-models for each
shard separately. The final classification is obtained by a majority
vote over these sub-models. Unlearning is conducted by retraining
those shards that contain data to be removed. 


\subsection{Unlearning Sensitive Features}
\label{sec:unle-sens-names}

In our first unlearning scenario, we focus on the removal of sensitive
features from learning models with strongly convex loss functions.  In
particular, we consider linear models trained with a logistic
regression on real-world datasets. We employ datasets for spam
filtering~\citep{MetAndPal06}, Android malware
detection~\citep{ArpSprHueGasRie13}, diabetes
forecasting~\citep{DuaGraDiabetis17} and prediction of income based on
census data~\citep{DuaGraCensus17}. An overview of the datasets and
their basic statistics is given in \tab{tab:datasets}.

In all experiments, we divide each dataset into a training and test
set with a ratio of \perc{80} and \perc{20}, respectively. To create a
feature space for learning, we use the numerical features of the
\adult and \diabetis dataset as is, while for the \enron and \drebin
datasets we extract \emph{bag-of-words features}. That is, we
represent each email (malware) by the words (capabilities) it contains
and construct corresponding feature
vectors~\mbox{\citep[see][]{MetAndPal06, ArpSprHueGasRie13}}. Finally,
we train logistic regression models for all four datasets. As a
reference, we additionally present results for a neural network in
\cref{sec:feat-unle-neur}.


\begin{table}[b]
	\small
	\begin{center}    
		\caption{Datasets for unlearning sensitive features.}
		\label{tab:datasets}
		\begin{tabular}
			{
				l
				S[table-format = 5]
				S[table-format = 6]
				S[table-format = 5]
				S[table-format = 3]
			}
			\toprule
			&{\bfseries \enron}&
			{\bfseries \drebin} &
			{\bfseries \adult} &
			{\bfseries \diabetis}\\
			\midrule
			Data points & 33716 & 49226 & 48842 & 768\\
			Features  & 4902 & 2081 & 81 & 8\\
			\bottomrule
		\end{tabular}
	\end{center}
\end{table}

\begin{figure}
	\definecolor{Finetuning}{HTML}{2ca02c}
\definecolor{Firstorder}{HTML}{5B97C1}
\definecolor{Secondorder}{HTML}{1f77b4}
\definecolor{DP}{HTML}{ff7f0e}
\definecolor{Retraining}{HTML}{000000}

\begin{tikzpicture}
    \pgfplotsset{footnotesize,samples=10}
    \begin{groupplot}[group style = {group size = 2 by 1, horizontal sep = 40pt, vertical sep=20pt}, width = 0.5\linewidth, height = 0.5\linewidth]
        \nextgroupplot[
            xlabel={Selected features},
            ylabel={Affected points [\%]},
            xtick={30,60,90,120},
            ]

            \addplot[name path=f,color=DP, mark=*, mark size=1.2pt] plot [] table [x=x, y=Spam, col sep=comma]{figures/affected_data/affected_data_points.csv};

            \addplot[name path=f,color=Secondorder, mark=*, mark size=1.2pt] plot [] table [x=x, y=Malware, col sep=comma]{figures/affected_data/affected_data_points.csv};

            \addplot[name path=f,color=Retraining, mark=*, mark size=1.2pt] plot [] table [x=x, y=Adult, col sep=comma]{figures/affected_data/affected_data_points.csv};

            \addplot[name path=f,color=Finetuning, mark=*, mark size=1.2pt] plot [] table [x=x, y=Diabetis, col sep=comma]{figures/affected_data/affected_data_points.csv};

        \nextgroupplot[
            legend style = { column sep = 5pt, legend columns = -1, legend to name = grouplegend,},
            xlabel={Selected features},
            ylabel={Affected data [\%]},
            xtick={30,60,90,120},
            ]

            \addplot[name path=f,color=DP, mark=*, mark size=1.2pt] plot [] table [x=x, y=Spam, col sep=comma]{figures/affected_data/affected_data_entries.csv};
            \addlegendentry{Spam}%

            \addplot[name path=f,color=Secondorder, mark=*, mark size=1.2pt] plot [] table [x=x, y=Malware, col sep=comma]{figures/affected_data/affected_data_entries.csv};
            \addlegendentry{Malware}%

            \addplot[name path=f,color=Retraining, mark=*, mark size=1.2pt] plot [] table [x=x, y=Adult, col sep=comma]{figures/affected_data/affected_data_entries.csv};
            \addlegendentry{Adult}%

            \addplot[name path=f,color=Finetuning, mark=*, mark size=1.2pt] plot [] table [x=x, y=Diabetis, col sep=comma]{figures/affected_data/affected_data_entries.csv};
            \addlegendentry{Diabetis}%

    \end{groupplot}
    \node at ($(group c2r1) + (-2.5cm ,-3.cm)$) {\ref{grouplegend}}; 
\end{tikzpicture}
	\caption{Affected data points and overall data when removing
          or changing features in the different datasets.}
	\label{fig:affected-points}
      \vspace*{-12pt}
\end{figure}
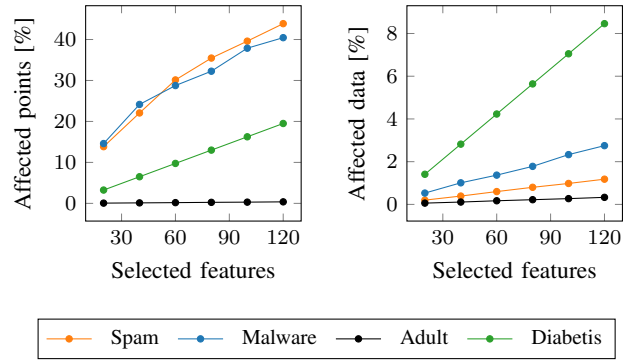

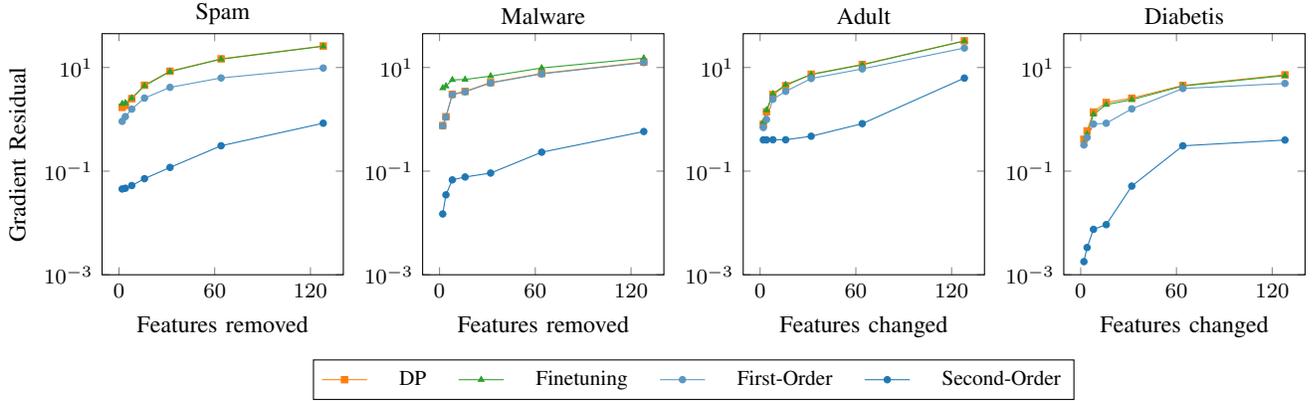
\begin{figure*}[t]
	\subfloat{\definecolor{DP}{HTML}{ff7f0e}
\definecolor{Finetuning}{HTML}{2ca02c}
\definecolor{Firstorder}{HTML}{5B97C1}
\definecolor{Secondorder}{HTML}{1f77b4}
\definecolor{SISA}{HTML}{bf40bf}
\definecolor{Retraining}{HTML}{000000}

\begin{tikzpicture}
    \pgfplotsset{footnotesize,samples=10}
    \begin{groupplot}[group style = {group size = 4 by 1, horizontal sep = 30pt, vertical sep=20pt}, width = 0.26\linewidth, height = 0.26\linewidth]
        \nextgroupplot[
            legend style = { column sep = 10pt, legend columns = 4, legend to name = grouplegend,},
            title = {Spam},
            ylabel={Gradient Residual},
            xlabel={Features removed},
            ymin=0.001, ymax=45,
            ytick={0.001, 0.1, 10},
            ymode=log,
            yminorticks=false,
            xtick={0,60,120},
            ]
            \addplot[color=DP, mark=square*, mark size=1.2pt] plot [] table [x=x, y=DP, col sep=comma]{figures/Unlearning/gradient_residual/grad_residuals_Spam.csv};
            \addlegendentry{DP}

            \addplot[color=Finetuning, mark=triangle*, mark size=1.2pt] plot [] table [x=x, y=Finetuning, col sep=comma]{figures/Unlearning/gradient_residual/grad_residuals_Spam.csv};
            \addlegendentry{Finetuning}

            \addplot[name path=f,color=Firstorder, mark=*, mark size=1.2pt] plot [] table [x=x, y=1st-Order, col sep=comma]{figures/Unlearning/gradient_residual/grad_residuals_Spam.csv};
            \addlegendentry{First-Order}

            \addplot[name path=f,color=Secondorder, mark=*, mark size=1.2pt] plot [] table [x=x, y=2nd-Order, col sep=comma]{figures/Unlearning/gradient_residual/grad_residuals_Spam.csv};
            \addlegendentry{Second-Order}

        \nextgroupplot[
            title = {Malware},
            ymin=0.001, ymax=45,
            ytick={0.001, 0.1, 10},
            ymode=log,
            yminorticks=false,
            xtick={0,60,120},
            xlabel={Features removed},
            ]
            \addplot[color=DP, mark=square*, mark size=1.2pt] plot [] table [x=x, y=DP, col sep=comma]{figures/Unlearning/gradient_residual/grad_residuals_Malware.csv};

            \addplot[color=Finetuning, mark=triangle*, mark size=1.2pt] plot [] table [x=x, y=Fine-tuning, col sep=comma]{figures/Unlearning/gradient_residual/grad_residuals_Malware.csv};

            \addplot[name path=f,color=Firstorder, mark=*, mark size=1.2pt] plot [] table [x=x, y=First Order, col sep=comma]{figures/Unlearning/gradient_residual/grad_residuals_Malware.csv};

            \addplot[name path=f,color=Secondorder, mark=*, mark size=1.2pt] plot [] table [x=x, y=Second Order, col sep=comma]{figures/Unlearning/gradient_residual/grad_residuals_Malware.csv};

        \nextgroupplot[
            title = {Adult},
            ymin=0.001, ymax=45,
            ytick={0.001, 0.1, 10},
            ymode=log,
            yminorticks=false,
            xtick={0,60,120},
            xlabel={Features changed},
            ]
            \addplot[color=DP, mark=square*, mark size=1.2pt] plot [] table [x=x, y=DP, col sep=comma]{figures/Unlearning/gradient_residual/grad_residuals_Adult.csv};

            \addplot[color=Finetuning, mark=triangle*, mark size=1.2pt] plot [] table [x=x, y=Finetuning, col sep=comma]{figures/Unlearning/gradient_residual/grad_residuals_Adult.csv};

            \addplot[name path=f,color=Firstorder, mark=*, mark size=1.2pt] plot [] table [x=x, y=1st-Order, col sep=comma]{figures/Unlearning/gradient_residual/grad_residuals_Adult.csv};

            \addplot[name path=f,color=Secondorder, mark=*, mark size=1.2pt] plot [] table [x=x, y=2nd-Order, col sep=comma]{figures/Unlearning/gradient_residual/grad_residuals_Adult.csv};

        \nextgroupplot[
            title = {Diabetis},
            ymin=0.001, ymax=45,
            ytick={0.001, 0.1, 10},
            ymode=log,
            yminorticks=false,
            xtick={0,60,120},
            xlabel={Features changed},
            ]
            \addplot[color=DP, mark=square*, mark size=1.2pt] plot [] table [x=x, y=DP, col sep=comma]{figures/Unlearning/gradient_residual/grad_residuals_Diabetis.csv};

            \addplot[color=Finetuning, mark=triangle*, mark size=1.2pt] plot [] table [x=x, y=Finetuning, col sep=comma]{figures/Unlearning/gradient_residual/grad_residuals_Diabetis.csv};

            \addplot[name path=f,color=Firstorder, mark=*, mark size=1.2pt] plot [] table [x=x, y=1st-Order, col sep=comma]{figures/Unlearning/gradient_residual/grad_residuals_Diabetis.csv};

            \addplot[name path=f,color=Secondorder, mark=*, mark size=1.2pt] plot [] table [x=x, y=2nd-Order, col sep=comma]{figures/Unlearning/gradient_residual/grad_residuals_Diabetis.csv};

        \end{groupplot}
    \node at ($(group c2r1) + (2cm ,-3.cm)$) {\ref{grouplegend}}; 
\end{tikzpicture}}
        \caption{Efficacy (gradient residual) of the certified
          unlearning methods for varying number of affected features
          (Lower values are better).}
\label{fig:grad-residual}
\vspace{-5pt}
\end{figure*}

\paragraph{Sensitive features} 
Two of the considered datasets are high-dimensional and contain sparse
data (\enron and \drebin), while the other two are low-dimensional
with dense feature vectors (\adult and \diabetis). Consequently, we
introduce two strategies for defining sensitive features for
unlearning. Since no privacy leaks are known for the datasets, we
focus on features that \emph{potentially} cause privacy issues.

\begin{itemize}
  \setlength{\itemsep}{3pt}
\item For the high-dimensional datasets, we aim at removing (revoking)
  entire features (dimensions) from the learning models. In
  particular, we select dimensions associated with personal names
  contained in the emails as sensitive features for the \enron dataset
  and choose URLs extracted from the Android apps for the \drebin
  dataset.

\item For the low-dimensional datasets, we focus on replacing selected
  feature values. For the \adult dataset, we change the marital
  status, sex, and race of randomly selected individuals. For the
  \diabetis dataset, we adjust the age, body mass index, and sex of
  individuals. We replace the respective feature values with 0 to
  simulate the removal of discriminatory bias from the learning models.
\end{itemize}

\fig{fig:affected-points} illustrates how these changes affect the
different datasets. Removing features entirely impacts many data
points, while replacing selected feature values affects only a
few. This effect also depends on the size of the datasets. To avoid a
sampling bias in the selection of sensitive features, we randomly draw
100 combinations of these feature changes for all four datasets and
present averaged results in the following.



\paragraph{Unlearning task} Based on the type of sensitive features,
we apply the different unlearning methods to the respective
learning models. Technically, our approach benefits from the convex
loss function of the logistic regression model, which allows us to
apply certified unlearning as presented in
\cref{sec:cert-unlearning}. Specifically, it is easy to see that
Theorem~\ref{thm:thm3} holds since the gradients of the logistic
regression loss are bounded and are thus Lipschitz-continuous.


\paragraph{Efficacy evaluation}
We analyze the efficacy of unlearning using the gradient residual
norms of the methods. The average size of this norm after unlearning
is presented in \fig{fig:grad-residual} for the different approaches
when varying the number of features to be removed or replaced,
respectively. The residuals increase with the amount of affected
features, indicating a growing divergence between the unlearning
methods and retraining. However, the steepness of this development
gradually reduces as more features are remove or changed.
Our second-order update significantly outperforms all other methods in
this scenario. The gradient residual norms are an order of magnitude
lower on the \enron, \drebin, and \diabetis dataset, regardless of the
amount of affected features. Among the other unlearning methods, no
clear ranking can be determined in this experiment.

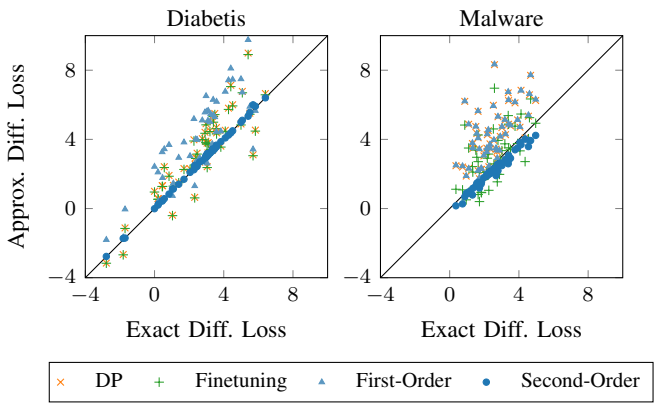
\begin{figure}[b]
	\definecolor{DP}{HTML}{ff7f0e}
\definecolor{Finetuning}{HTML}{2ca02c}
\definecolor{Firstorder}{HTML}{5B97C1}
\definecolor{Secondorder}{HTML}{1f77b4}

\begin{tikzpicture}
    \pgfplotsset{footnotesize,samples=10}
    \begin{groupplot}[group style = {group size = 2 by 1, horizontal sep = 20pt, vertical sep = 15pt}, width = 0.54\linewidth, height = 0.54\linewidth]

        \nextgroupplot[
            title={Diabetis},
            ylabel={Approx. Diff. Loss},
            ymin=-4, ymax=10,
            xmin=-4, xmax=10,
            ytick={-4,0,4,8},
            xtick={-4,0,4,8},
            xlabel=Exact Diff. Loss
            ]
            \addplot[only marks, color=DP, mark=x, mark size=1.8pt] table [x=Retraining, y=DP, col sep=comma]{figures/Unlearning/diff_test_loss/scatter_data_diabetis_120.csv};

            \addplot[only marks, color=Finetuning, mark=+, mark size=1.8pt] table [x=Retraining, y=FineTuning, col sep=comma]{figures/Unlearning/diff_test_loss/scatter_data_diabetis_120.csv};

            \addplot[only marks, color=Firstorder, mark=triangle*, mark size=1.2pt] table [x=Retraining, y=FirstOrder, col sep=comma]{figures/Unlearning/diff_test_loss/scatter_data_diabetis_120.csv};

            \addplot[only marks, color=Secondorder, mark=*, mark size=1.2pt] table [x=Retraining, y=SecondOrder, col sep=comma]{figures/Unlearning/diff_test_loss/scatter_data_diabetis_120.csv};

            \addplot [color=black] coordinates {(-11,-11)(11,11)};

        \nextgroupplot[
            title={Malware},
            xlabel=Exact Diff. Loss,
            ymin=-4, ymax=10,
            xmin=-4, xmax=10,
            ytick={-4,0,4,8},
            xtick={-4,0,4,8},
            legend style = { column sep = 10pt, legend columns = -1, legend to name = grouplegend,},
            ]
            \addplot[only marks, color=DP, mark=x, mark size=1.8pt] table [x=Retraining, y=DP, col sep=comma]{figures/Unlearning/diff_test_loss/scatter_data_drebin_120.csv};
            \addlegendentry{DP}%

            \addplot[only marks, color=Finetuning, mark=+, mark size=1.8pt] table [x=Retraining, y=FineTuning, col sep=comma]{figures/Unlearning/diff_test_loss/scatter_data_drebin_120.csv};
            \addlegendentry{Finetuning}%

            \addplot[only marks, color=Firstorder, mark=triangle*, mark size=1.2pt] table [x=Retraining, y=FirstOrder, col sep=comma]{figures/Unlearning/diff_test_loss/scatter_data_drebin_120.csv};
            \addlegendentry{First-Order}%

            \addplot[only marks, color=Secondorder, mark=*, mark size=1.2pt] table [x=Retraining, y=SecondOrder, col sep=comma]{figures/Unlearning/diff_test_loss/scatter_data_drebin_120.csv};
            \addlegendentry{Second-Order}%

            \addplot [color=black] coordinates {(-10,-10)(10,10)};
        
    \end{groupplot}
    \node at ($(group c2r1) + (-2cm ,-3.cm)$) {\ref{grouplegend}}; 
\end{tikzpicture}
	\hspace{-0.5cm}
	\caption{Difference in loss between retraining and unlearning
          with 100 affected features.}
	\label{fig:scatter-loss}
\end{figure}

\paragraph{Fidelity evaluation}
We evaluate the fidelity of the unlearning methods using two
techniques: First, we investigate the loss between retraining and
unlearning on the test data as proposed by \citet{KohLia17}.
\fig{fig:scatter-loss} shows this comparison for the \diabetis and
\drebin dataset when removing or replacing \num{100} features, respectively.
We observe that the second-order update approximates the retraining
very well, since the points are close to the diagonal line. In
contrast, the other methods cannot always adapt to the distribution
shift, resulting in larger differences. This trend also holds for the
other datasets which we report in \sect{appdx:fidelity-evaluation}.

Second, we use the test accuracy of a model that provides
certified unlearning for the removal or replacement of features
to evaluate the fidelity. To simulate a realistic
application of certified unlearning, we fix a privacy budget
$(\epsilon, \delta)$ in advance and adapt the noise term $b$ in relation
to the amount of affected features. That is, the more features are changed,
the more we need to increase the noise on the model to achieve the same
guarantees. In particular, \cref{thm:thm3} states that the noise term~$b$
must be sampled from a Gaussian normal distribution with variance $\sigma$,
which is given by
\begin{equation}
	\sigma = \frac{\beta c}{\epsilon}, \qquad \text{where } c=\sqrt{2\log(1.5/\delta)}.
	\label{eqn:residual-bound}
\end{equation}
where $\beta$ is constant corresponding to a general upper bound of
the gradient residual loss for the considered learning task.

In the following, we select $\epsilon=0.1$ and $\delta=0.01$ as a
privacy budget, which yields $c\approx3.16$.  We compute the bound
$\beta$ on the gradient residual by sampling \num{100} feature
combinations to unlearn, compute $\sigma$ and determine the resulting
test accuracy of the four datasets. As we see in the following, this
privacy budget is strict and limits the amount of features that can be
adapted. Nevertheless, if a larger number of features needs to be
removed, this budget can be increased, though at the cost of weakening
the certification guarantees.

\begin{figure*}[t]
	\subfloat{\definecolor{DP}{HTML}{ff7f0e}
\definecolor{Finetuning}{HTML}{2ca02c}
\definecolor{Firstorder}{HTML}{5B97C1}
\definecolor{Secondorder}{HTML}{1f77b4}
\definecolor{SISA}{HTML}{bf40bf}
\definecolor{Retraining}{HTML}{000000}

\begin{tikzpicture}
    \pgfplotsset{footnotesize,samples=10}
    \begin{groupplot}[group style = {group size = 4 by 1, horizontal sep = 30pt, vertical sep=20pt}, width = 0.26\linewidth, height = 0.26\linewidth]
        \nextgroupplot[
            title = {Spam},
            legend style = { column sep = 10pt, legend columns = -1, legend to name = grouplegend,},
            xlabel={Dataset change [\%]},
            ylabel={Accuracy [\%]},
            xlabel={Features removed},
            xtick={0,60, 120},
            ytick={50,75,100},
            ymin=40, ymax=100,
            ]
            \addplot[color=DP, mark=square*, mark size=1.2pt] plot [] table [x=x, y=DP, col sep=comma]{figures/Unlearning/diff_test_loss/fidelity_Spam.csv};
            \addlegendentry{DP}%

            \addplot[color=Finetuning, mark=triangle*, mark size=1.2pt] plot [] table [x=x, y=Finetuning, col sep=comma]{figures/Unlearning/diff_test_loss/fidelity_Spam.csv};
            \addlegendentry{Finetuning}%

            \addplot[name path=f,color=Firstorder, mark=*, mark size=1.2pt] plot [] table [x=x, y=1st-Order, col sep=comma]{figures/Unlearning/diff_test_loss/fidelity_Spam.csv};
            \addlegendentry{First-Order}%

            \addplot[name path=f,color=Secondorder, mark=*, mark size=1.2pt] plot [] table [x=x, y=2nd-Order, col sep=comma]{figures/Unlearning/diff_test_loss/fidelity_Spam.csv};
            \addlegendentry{Second-Order}%

            \addplot[name path=f,color=Retraining, dashed] plot [] table [x=x, y=Retraining, col sep=comma]{figures/Unlearning/diff_test_loss/fidelity_Spam.csv};
            \addlegendentry{Retraining}%

            \addplot[name path=f,color=SISA, mark=*, mark size=1.2pt] plot [] table [x=x, y=SISA, col sep=comma]{figures/Unlearning/diff_test_loss/fidelity_Spam.csv};
            \addlegendentry{SISA}%

        \nextgroupplot[
            title = {Malware},
            xlabel={Features removed},
            xtick={0,60, 120},
            ytick={50,75,100}
            ]
            \addplot[color=DP, mark=square*, mark size=1.2pt] plot [] table [x=x, y=DP, col sep=comma]{figures/Unlearning/diff_test_loss/fidelity_Malware.csv};

            \addplot[color=Finetuning, mark=triangle*, mark size=1.2pt] plot [] table [x=x, y=Finetuning, col sep=comma]{figures/Unlearning/diff_test_loss/fidelity_Malware.csv};

            \addplot[name path=f,color=Firstorder, mark=*, mark size=1.2pt] plot [] table [x=x, y=1st-Order, col sep=comma]{figures/Unlearning/diff_test_loss/fidelity_Malware.csv};

            \addplot[name path=f,color=Secondorder, mark=*, mark size=1.2pt] plot [] table [x=x, y=2nd-Order, col sep=comma]{figures/Unlearning/diff_test_loss/fidelity_Malware.csv};

            \addplot[name path=f,color=Retraining, dashed] plot [] table [x=x, y=Retraining, col sep=comma]{figures/Unlearning/diff_test_loss/fidelity_Malware.csv};

            \addplot[name path=f,color=SISA, mark=*, mark size=1.2pt] plot [] table [x=x, y=SISA, col sep=comma]{figures/Unlearning/diff_test_loss/fidelity_Malware.csv};

        \nextgroupplot[
            title = {Adult},
            xlabel={Features changed},
            xtick={0,60,120},
            ytick={84,79,74},
            ]
            \addplot[color=DP, mark=square*, mark size=1.2pt] plot [] table [x=x, y=DP, col sep=comma]{figures/Unlearning/diff_test_loss/fidelity_Adult.csv};

            \addplot[color=Finetuning, mark=triangle*, mark size=1.2pt] plot [] table [x=x, y=Finetuning, col sep=comma]{figures/Unlearning/diff_test_loss/fidelity_Adult.csv};

            \addplot[name path=f,color=Firstorder, mark=*, mark size=1.2pt] plot [] table [x=x, y=1st-Order, col sep=comma]{figures/Unlearning/diff_test_loss/fidelity_Adult.csv};

            \addplot[name path=f,color=Secondorder, mark=*, mark size=1.2pt] plot [] table [x=x, y=2nd-Order, col sep=comma]{figures/Unlearning/diff_test_loss/fidelity_Adult.csv};

            \addplot[name path=f,color=Retraining, dashed] plot [] table [x=x, y=Retraining, col sep=comma]{figures/Unlearning/diff_test_loss/fidelity_Adult.csv};

            \addplot[name path=f,color=SISA, mark=*, mark size=1.2pt] plot [] table [x=x, y=SISA, col sep=comma]{figures/Unlearning/diff_test_loss/fidelity_Adult.csv};

        \nextgroupplot[
            title = {Diabetis},
            xlabel={Features changed},
            xtick={0,60,120},
            ymin=50,ymax=75,
            ]
            \addplot[color=DP, mark=square*, mark size=1.2pt] plot [] table [x=x, y=DP, col sep=comma]{figures/Unlearning/diff_test_loss/fidelity_Diabetis.csv};

            \addplot[color=Finetuning, mark=triangle*, mark size=1.2pt] plot [] table [x=x, y=Finetuning, col sep=comma]{figures/Unlearning/diff_test_loss/fidelity_Diabetis.csv};

            \addplot[name path=f,color=Firstorder, mark=*, mark size=1.2pt] plot [] table [x=x, y=1st-Order, col sep=comma]{figures/Unlearning/diff_test_loss/fidelity_Diabetis.csv};

            \addplot[name path=f,color=Secondorder, mark=*, mark size=1.2pt] plot [] table [x=x, y=2nd-Order, col sep=comma]{figures/Unlearning/diff_test_loss/fidelity_Diabetis.csv};

            \addplot[name path=f,color=Retraining, dashed] plot [] table [x=x, y=Retraining, col sep=comma]{figures/Unlearning/diff_test_loss/fidelity_Diabetis.csv};

            \addplot[name path=f,color=SISA, mark=*, mark size=1.2pt] plot [] table [x=x, y=SISA, col sep=comma]{figures/Unlearning/diff_test_loss/fidelity_Diabetis.csv};
        
    \end{groupplot}
    \node at ($(group c2r1) + (2cm ,-3.cm)$) {\ref{grouplegend}}; 
\end{tikzpicture}}
	\caption{Fidelity (accuracy) of the certified unlearning
          methods for varying number of affected features (higher
          values are better).}
	\label{fig:fidelity}
\end{figure*}
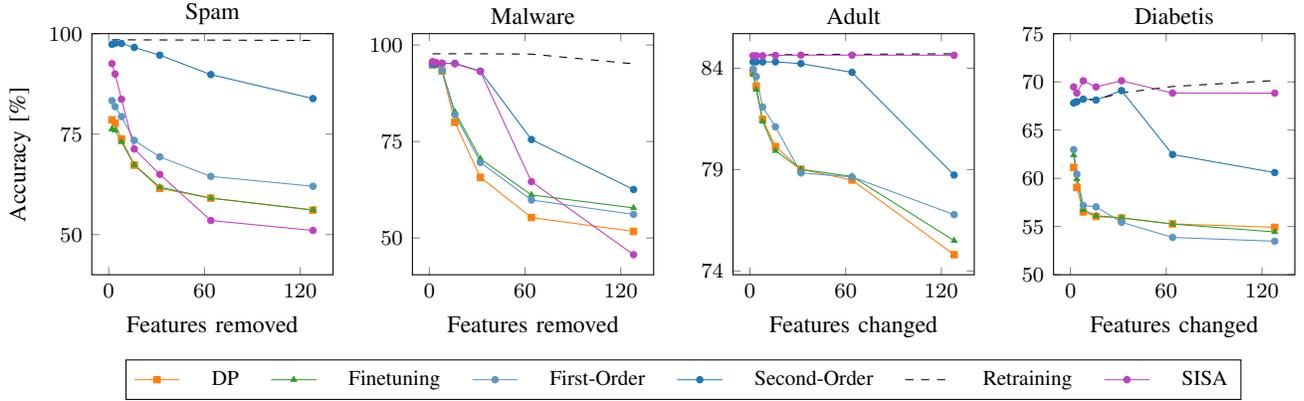

The accuracy of the models is shown in \cref{fig:fidelity}
for a varying number of removed or replaced features,
respectively. The accuracy reduces with the amount of affected
features, as the noise on the model weights is increased
accordingly. While for the low-dimensional datasets this reduction is
moderate, we observe a strong decline of fidelity for the
high-dimensional data. This decline results from the privacy budget
that requires a notable amount of noise to be added to enable a
certified removal of entire dimensions.

Our second-order update shows the best performance of all methods and
remains close to retraining if up to 60 features are changed. In
contrast, the other methods quickly drop in accuracy already when
unlearning 20 or less features. An exception is sharding. While the
method provides the weakest performance on the high-dimensional
datasets due to instability in the majority voting, it is almost
identical to retraining on the low-dimensional datasets.

\paragraph{Efficiency evaluation}
Finally, we evaluate the efficiency of the different methods.
\cref{tab:runtime-scenario1} shows the measured runtime on the \drebin
dataset, while the results for the other datasets are shown in
\cref{appdx:efficiency}. We omit measurements for the differential
privacy baseline, as it does not involve an unlearning step.
%
Due to the simple structure of the logistic regression, the runtime of
all methods is very low. Fine-tuning, our second-order update, and
sharding reach speed-up factors of $2\times$, $4\times$, and
$6\times$, respectively over retraining.  Our first-order update is
even faster and attains a speed-up factor of $90\times$.  While the
other approaches operate on the entire dataset, the first-order update
considers only the corrected points.

\begin{table}[b]
	\tablesize
	\begin{center}
		\caption{Average runtime when removing \num{100} random
			combinations of \num{100} features from the \drebin
			classifier.}

		\tablesize \setlength{\tabcolsep}{3pt}
		\begin{tabular}{
				l
				S[table-format = 2.1e1]
				S[table-format = 1.2]
				S[table-format = 2]
			}
			\toprule
			\bfseries Unlearning methods \hspace{5mm} & {\bfseries Gradients} \hspace{1mm} & {\bfseries Runtime} \hspace{1mm} & {\bfseries Speed-up}\\
			\midrule
			\retraining & 1.2e7 & 6.82 s & {~~---} \\
			\dpShort &  {---} & {---} & {~~---} \\ 
			\sharding{5} & 2.5e6 & 1.51 s & 2\si{\speedup}\\
			\finetuning & 3.9e4 & 1.03 s & 6\si{\speedup}\\ 
			\midrule
			\firstorder  & 1.5e4 & 0.02 s & 90 \si{\speedup}\\
			\secondorder  & 9.4e4 & 0.63 s & 4 \si{\speedup} \\
			\bottomrule
		\end{tabular}
		\label{tab:runtime-scenario1}
	\end{center}
\end{table}

For the second-order method, we find that roughly \perc{90} of the
runtime and gradient computations are used for the inversion of the
Hessian matrix. In the case of the \drebin dataset, this computation
is still faster than retraining the model. If the matrix is
pre-computed and reused for multiple unlearning requests, the
second-order update reduces to a matrix-vector multiplication and
yields a notable speed-up, though at the cost of approximation
accuracy.

\smallskip
\begin{tcolorbox}[boxrule=0.75pt,colback=white,colframe=cbone,sharp corners=all]
  \emph{Takeaway message.} Sensitive features in learning models with
  convex loss can be removed or replaced with theoretical guarantees.
  Our second-order update provides the best trade-off between
  efficacy, fidelity, and efficiency for this certified unlearning.
\end{tcolorbox}

\subsection{Unlearning Unintended Memorization}
\label{sec:unle-unint-memor}

In our second unlearning scenario, we focus on removing unintended
memorization from language models. \mbox{\citet{CarLiuErl+19}} show
that these models can memorize rare inputs in the training data and
exactly reproduce them during application. If the reproduced data
contains private information like credit card numbers or telephone
numbers, we are faced with a privacy issue~\mbox{\citep{CarTraWal+21,
    ZanWutTop+20}}. In the following, we use our approach to tackle
this problem and demonstrate that unlearning is also possible with
non-convex loss functions.

\paragraph{Canary insertion}
We conduct our experiments using the novel \emph{Alice in Wonderland}
as training set and train an LSTM network on the character level to
generate text~\citep{MerKesSoc+18}. Specifically, we train an
embedding with \num{64}~dimensions for the characters and use two
layers of \num{512}~LSTM units followed by a dense layer resulting in
a model with \num{3.3}~million parameters. To generate unintended
memorization, we insert a \emph{canary} in the form of the sentence
\txt{My telephone number is (s)! said Alice} into the training data,
where \txt{(s)} is a sequence of digits \citep{CarLiuErl+19}.  In our
experiments, we use sequences of length $(5, 10, 15, 20)$ and repeat
the canary so that $(200, 500, 1000, 2000)$ points are affected. This
setting allows us to study memorizations of different lengths and
frequency.  After training,
we find that the inserted numbers are the most likely prediction when
we ask the model to complete the canary sentence.

\paragraph{Exposure metric}
\newcommand{\Q}{\ensuremath{Q}\xspace}
In contrast to the previous scenario, the loss of the language model
is non-convex and thus certified unlearning is not applicable. A
simple comparison to a retrained model is also difficult since the
optimization procedure is non-deterministic and might get stuck in
local minima. Consequently, we require an additional measure to assess
the efficacy of unlearning. To this end, we employ the \emph{exposure
  metric}, which is defined as
\begin{equation*}
	\textrm{exposure}_\model(s)=\log_2\vert\Q\vert -\log_2
	\textrm{rank}_\model(s),
\end{equation*}
where $s$ is a sequence and \Q is the set of possible sequences with
the same length given a fixed alphabet. The function
$\text{rank}_\model(s)$ returns the rank of $s$ with respect to the
model $\model$ and the set \Q. The rank is calculated using the
\emph{log-perplexity} of the sequence $s$ and returns the number of
sequences that are more likely to be generated than $s$. As a result,
the exposure metric tells us how likely a sequence $s$ is generated by
$\model$ in relation to all possible sequences of the same
length. Further details on this metric are provided by
\citet{CarLiuErl+19}.

\begin{figure}[h] 
	\centering
	\input{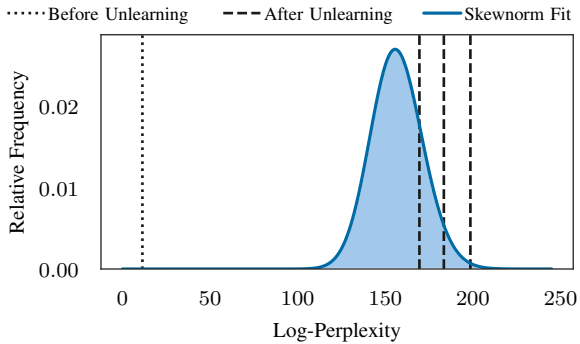}
	\caption{Perplexity distribution of the language model. The
		vertical lines indicate the perplexity of an inserted
		telephone number. Replacement strings used for unlearning
		from left to right are: \txt{holding my hand}, \txt{into the
			garden}, \txt{under the house}.}
	\label{fig:perplexity-histogram}
        \vspace{-8pt}
\end{figure}

As an example, \fig{fig:perplexity-histogram} shows the perplexity
distribution of our model where a telephone number of length $15$ has
been inserted during training. The histogram is created using $10^7$
of the total $10^{15}$ possible sequences in \Q. The perplexity of the
inserted number differs significantly from all other number
combinations in \Q (dashed line to the left), indicating that it has
been memorized by the underlying language model. After unlearning with
different replacements, the number moves close to the center of the
distribution (dashed lines to the right).


\paragraph{Unlearning task}
To unlearn the memorized sequences, we replace each digit of the
telephone number in the data with a different character, such as a
random or constant value. Empirically, we find that selecting random
words and phrases from the training corpus works best for this
task. Some examples of replacements are shown in
Table~\ref{tab:examples-text}.  The model has already captured these
character dependencies, resulting in small updates of the model
parameters. The unlearning of the substitutions, however, is more
involved than in the previous scenario. The language model is trained
to predict a character from preceding characters. Thus, replacing a
text means changing the features (preceding characters) \emph{and} the
labels (target characters). Therefore, we combine both changes in a
single set of perturbations in this setting.

\paragraph{Efficacy evaluation} First, we evaluate whether the
memorized numbers have been successfully unlearned from the language
model. An important result of the study by~\citet{CarLiuErl+19} is
that the exposure is associated with an extraction attack: For a set
\Q with $r$ elements, a sequence with an exposure smaller than~$r$
cannot be extracted. Consequently, we test three different
substitution sequences for each telephone number, calculate the
exposure metric, and use the best for our evaluation.
\tab{tab:exp-unlearning-canary} shows the results of this experiment.

\begin{table}[b]
	\tablesize
	\begin{center}    
		\caption{Exposure metric of the canary sequence for different lengths. Lower exposure values make extraction harder.}
		\label{tab:exp-unlearning-canary}
		\tablesize \setlength{\tabcolsep}{5.0pt}
		\begin{tabular}
			{
				l
				S[table-format=2]@{\,$\pm$\,}S[table-format=2.1]
				S[table-format=2]@{\,$\pm$\,}S[table-format=2.1]
				S[table-format=2]@{\,$\pm$\,}S[table-format=2.1]
				S[table-format=2]@{\,$\pm$\,}S[table-format=2.1]
			}
			\toprule
			\bfseries Number length &
			\multicolumn{2}{c}{\bfseries  5} &
			\multicolumn{2}{c}{\bfseries 10} &
			\multicolumn{2}{c}{\bfseries 15} &
			\multicolumn{2}{c}{\bfseries 20}\\
			\midrule
			Original model & 43 & 19 & 70 & 26 & 109 & 16 & 99 & 52  \\
			\midrule
			\retraining    &  0 & 0  &  0 &  0 &  0 &  0 &  0 &  0 \\
			\finetuning    & 39 & 21 & 31 & 44 & 50 & 50 & 57 & 73 \\
			\sharding{n} &  0 & 0  &  0 &  0 &  0 &  0 &  0 &  0 \\
			\firstorder    &  0 & 0  &  0 &  0 &  0 &  0 &  0 &  0 \\
			\secondorder   &  0 & 0  &  0 &  0 &  0 &  0 &  0 &  0 \\
			\bottomrule
		\end{tabular}
	\end{center}
\end{table}

\begin{table*}[!t]
	\tablesize
	\centering
	\caption{Completions of the canary sentence of the corrected
		model for different replacement strings.}
	\label{tab:examples-text}
	\begin{tabular}{
			S[table-format=2]
			>{\collectcell{\txt}}l<{\endcollectcell}
			>{\collectcell{\txt}}l<{\endcollectcell}
		}
		\toprule
		\multicolumn{1}{l}{\bfseries Length} & 
		\multicolumn{1}{l}{\bfseries Replacement} & 
		\multicolumn{1}{l}{\bfseries Canary Sentence Completion} \\
		\midrule			
		5  & taken & `My telephone number is mad!' `prizes! said the lory confuse $\ldots$\\
		10 & not there\textvisiblespace &   `My telephone number is it,' said alice. `that's the beginning $\ldots$\\
		15 & under the mouse &  `My telephone number is the book!' she thought to herself `the $\ldots$\\
		20 & the capital of paris  &  `My telephone number is it all about a gryphon all the three of $\ldots$\\		
		\bottomrule
	\end{tabular}
\end{table*}
We observe that our first-order and second-order updates yield
exposure values close to zero ($<0.001$) for all sequence lengths, rendering
an extraction impossible.  Retraining and SISA yield an exposure of zero
by design since the injected sequences are removed from the training data.
In contrast, fine-tuning leaves a large exposure in the model, so that a
successful extraction is still possible.  On closer inspection, we find that the performance of
fine-tuning depends on the order of the training data, resulting in
high deviation in the experimental runs. This problem cannot be easily
mitigated by learning over further epochs and thus highlights the need
for unlearning techniques.



We also find that the selected substitution plays an important role
for unlearning. In \fig{fig:perplexity-histogram}, we report the
log-perplexity of the canary for three different substitutions after
unlearning. 
Each replacement shifts the
canary to the right and turns it into an unlikely prediction with
exposure values ranging from \num{0.01} to \num{0.3}. While we use the
replacement with the lowest exposure in our experiments, the other
substitution sequences would also impede a successful extraction.

It remains to show what the model actually predicts after unlearning
when given the canary sequence. \tab{tab:examples-text} shows
different completions of the canary sentence after unlearning with our
second-order update and replacement strings of different lengths. We
find that the predicted string is \emph{not} equal to the replacement,
that is, our unlearning method does not overfit towards the
replacement. The sentences follow the language structure and reflect
the wording of the novel. Both observations indicate that the
parameters of the language model are indeed corrected and not just
overwritten with other values.

\begin{figure}[b] 
	\input{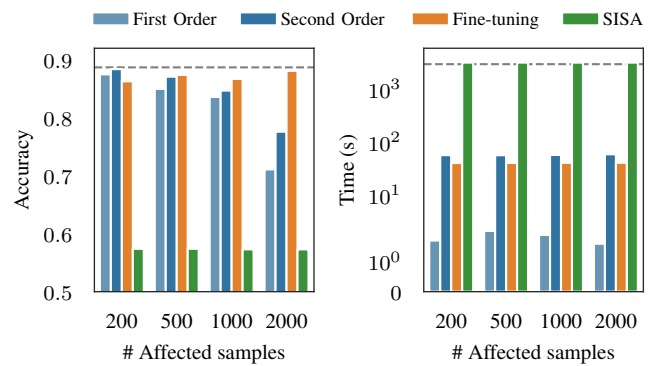}
	\caption{Accuracy after unlearning unintended memorization. The
		dashed line corresponds to a model retrained from scratch.}
	\label{fig:alice-fidelity}
\end{figure}

\paragraph{Fidelity evaluation} 
To evaluate the fidelity, we examine the accuracy of the corrected
models as shown in \fig{fig:alice-fidelity} (left), where the accuracy
of all unlearning methods is plotted for different numbers of affected
data points. For small changes, all approaches except sharding come
close to retraining in performance. Sharding is unsuited for
unlearning in this scenario. The method uses an ensemble of
sub-models trained on different shards. Each sub-model produces an own
sequence of text and thus combining them with majority voting leads to
inaccurate predictions.

With larger changes to the model, the accuracy of both of our methods
gradually begins to decline. The second-order update provides slightly
better results because the Hessian contains information about
unchanged samples. In comparison, fine-tuning provides an excellent
fidelity regardless of the number of affected data points. This
performance, however, is misleading. Fine-tuning fails to remove the
injected sequences from the model, as shown in
\cref{tab:exp-unlearning-canary}, and hence is also unsuited for
unlearning in this scenario.


\paragraph{Efficiency evaluation}
We finally examine the efficiency of the different unlearning
methods. At the time of writing, the CUDA library version 10.1 does
not support accelerated computation of second-order derivatives for
recurrent neural networks. Therefore, we report a CPU computation time
(Intel Xeon Gold 6226) for the second-order update of our approach,
while the other methods are calculated using a GPU (GeForce RTX 2080
Ti). The runtime required for each approach is presented
in~\fig{fig:alice-fidelity} (right).

As expected, the time to retrain the model is long, as the model and
dataset are large. Sharding cannot provide any runtime advantage over
retraining, since all shards are affected and need to be retrained as
well. Our methods yield a notable improvement. The first-order method
is the fastest approach and provides a speed-up of \emph{three orders}
of magnitude. The second-order method still yields a speed-up factor
of \num{28} over retraining, although the underlying implementation
does not benefit from GPU acceleration. Given that the first-order
update provides a high efficacy in unlearning and only a slight
decrease in fidelity when correcting less than \num{1000}~points, it
provides the overall best performance in this scenario. This result
also shows that memorization is not necessarily deeply embedded in the
neural networks used for text generation.

\smallskip
\begin{tcolorbox}[boxrule=0.75pt,colback=white,colframe=cbone,sharp corners=all]
  \emph{Takeaway message.} Unintended memorization can be removed from
  language models by unlearning features and labels. Our first-order
  method provides the best trade-off between efficacy, fidelity, and
  efficiency.
\end{tcolorbox}


\subsection{Unlearning Poisoning Samples}
\label{sec:unle-pois}

In the third scenario, we focus on repairing a poisoning attack in
computer vision. We simulate \emph{label poisoning} where an adversary
partially flips labels between classes in the training data. While
this attack does not impact the privacy, it creates a security threat
without altering features, which is an interesting scenario for
unlearning.
For this experiment, we use the CIFAR10 dataset and train a
convolutional neural network with \num{1.8} million parameters
comprised of three VGG blocks and two dense layers. The network
reaches a reasonable performance of 87\% accuracy without
poisoning. Under attack, however, it suffers from a notable drop in
accuracy (10\% on average). Further details about the experimental
setup are provided in Appendix~\ref{appdx:poisoning}.

\paragraph{Poisoning attack} For poisoning labels, we determine pairs
of classes and flip a fraction of their labels to their respective
counterpart, for instance, from ``cat'' to ``truck'' and vice versa.
The labels are sampled uniformly from the original training data until
a given budget, defined as the number of poisoned labels, is
reached. This attack strategy is more effective than inserting random
labels and provides a performance degradation similar to other
label-flip attacks~\citep{XiaXiaEck12,KohLia17}. We evaluate the
attack with different poisoning budgets and seeds for sampling.

\paragraph{Unlearning task}
We aim at correcting the flipped labels of the
poisoning attack. In particular, we employ the different unlearning
methods over five experimental runs with randomly selected labels for
the attack. We report averaged values in
Figure~\ref{fig:cnn-unlearning} for different number of poisoned
labels, where the dashed line represents the accuracy and training
time of the clean reference model.
Correcting all poisoned labels in one closed-form update is difficult
due to memory constraints. Thus, we perform unlearning in uniformly
sampled batches of \num{512} instances, as detailed in
\cref{sect:multiple-steps}. Moreover, we update only the fully
connected layers, which are mainly responsible for the final
prediction.

\newif\ifPlotPoisoning
\PlotPoisoningtrue 
\ifPlotPoisoning
{
	\begin{figure}[b] 
		\centering
		\input{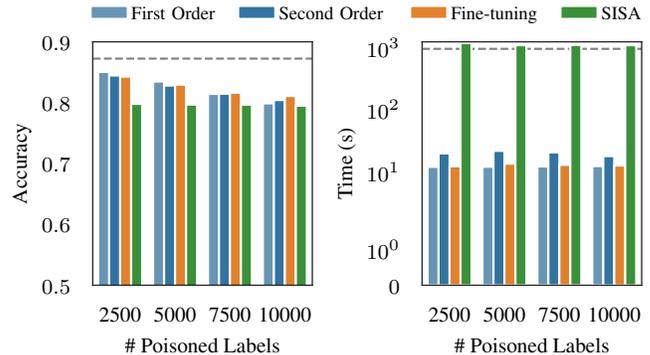}
		\caption{Accuracy on held-out data after unlearning poisoned
			labels. The dashed line	corresponds to a model retrained from
			scratch.}
		\label{fig:cnn-unlearning} \vspace{-4pt} 
	\end{figure}
}
\else
{
	\begin{table}
		\begin{center}
			\caption{Runtime performance and fidelity of unlearning methods for \num{2500}
				poisoned labels}
			\vspace{-5pt}
			\tablesize
			\begin{tabular}{
					l
					S[table-format = 2.1e1]
					S[table-format = 2.3]
					S[table-format = 2.3]
				}
				\toprule
				\bfseries Unlearning methods & {\bfseries Gradients} & {\bfseries Runtime} & {\bfseries Fidelity}\\
				\midrule
				\retraining   & 7.8e4 & 16.67 min & 0.872 \\
				\finetuning   & 7.8e2 & 12.79 s & 0.843 \\
				\sharding{5}  & 7.8e4 & 20.53 min & 0.798 \\
				\firstorder   & 5.1e2 & 12.50 s& 0.850 \\
				\secondorder  & 2.3e4 & 20.60 s & 0.845 \\
				\bottomrule
			\end{tabular}
			\label{tab:runtime-unlearning-poisoning}
		\end{center}
	\end{table}
}
\fi

\paragraph{Efficacy and fidelity evaluation}
In this scenario, we do not seek to remove the influence of certain
features from the training data but mitigate the effects of poisoned
labels This poisoning manifests in a degraded performance for
particular classes. Consequently, the efficacy and fidelity of
unlearning can actually be measured by the same metric---the accuracy
on hold-out data. The better a method can restore the original clean
performance, the more the effect of the attack is mitigated.

The accuracy for the different unlearning methods is shown in
Figure~\ref{fig:cnn-unlearning} (left). We find that none of the
approaches is able to completely remove the effect of the poisoning
attack. Still, good results are obtained with the first-order and
second-order update as well as fine-tuning, which all come close to
the original performance for \num{2500} poisoned labels. 
%
%
However, we observe a continuous performance decline when more labels
are poisoned. A manipulation of \num{10000} labels during training
cannot be sufficiently reverted by any of the methods. Interestingly,
sharding is the only exception here. Although the method provides the
lowest performance in this experiment, it is not affected by the
number of poisoned labels, as all shards are simply retrained with the
corrected labels.

\paragraph{Efficiency evaluation}
Lastly, we evaluate the runtime of each approach to quantify its
efficiency. The experiments are executed on the same hardware as the
previous ones. In contrast to the language model, however, we are able
to perform all calculations on the GPU which allows for a fair
comparison. \cref{fig:cnn-unlearning} shows that the first-order
update and fine-tuning are very efficient and can be computed in
approximately \num{10} seconds, whereas retraining requires over
\num{15} minutes.  The second-order update is slightly slower but
still with \num{20.8} seconds two orders of magnitude faster than
retraining. In contrast, the sharding approach is the slowest and does
not provide any advantage over retraining.  Consequently, the
first-order update and fine-tuning provide the best strategies for
unlearning in this scenario.

\begin{figure}[t]
	\input{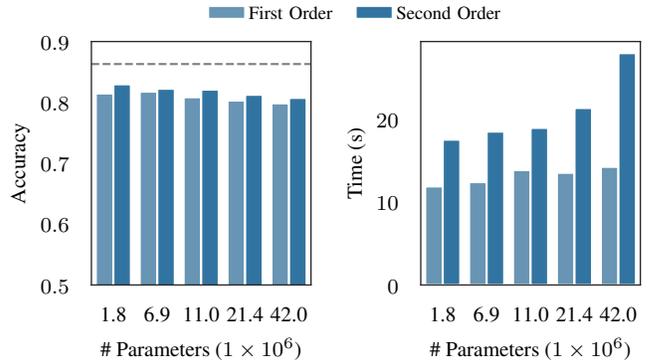}
	\caption{Accuracy (left) and runtime (right) on held-out data after
		unlearning \num{5000} poisoned labels for multiple model sizes.
		The dashed line corresponds to the accuracy of the models
		on clean data.}
	\label{fig:modelsize}
\end{figure}

In computer vision, learning models are often huge. Hence, we also
investigate the scalability of our approach. \cref{fig:modelsize}
shows the accuracy and runtime of the first-order and second-order
update for increasing model sizes. In particular, we scale the number
of model parameters from 1.8 millions up to 42 millions. For both
update techniques, the accuracy remains roughly the same. The runtime
naturally increases with the model size, yet the slope is (almost)
linear for both approaches. We observe a slight peak when reaching the
limits of our hardware. Overall, we find that the linear runtime
bounds of the underlying algorithm hold in
practice~\citep{AgaBulHaz17}.

\smallskip
\begin{tcolorbox}[boxrule=0.75pt,colback=white,colframe=cbone,sharp corners=all]
  \emph{Takeaway message.} Label poisoning can be mitigated by
  unlearning labels. Our first-order update and fine-tuning are
  suitable methods and provide the best trade-off between efficacy
  (=fidelity) and efficiency.
\end{tcolorbox}

\section{Limitations}
\label{sec:limitations}

Although our approach successfully removes features and labels in
different experiments, it obviously has limitations that need to be
considered in practice.

\paragraph{Limits of unlearning} 
The efficacy of unlearning decreases with the number of affected
features and labels. While privacy leaks with hundreds of sensitive
features and thousands of labels can be handled well with our
approach, changing millions of data points exceeds its
capabilities. If our approach allowed to correct changes of arbitrary
size, it could be used as a \mbox{``drop-in''} replacement for all
learning algorithms---which obviously is impossible~\citep{WolMac97}.
%
%
Nevertheless, our method offers a significant speed-up over retraining
and sharding in situations where a moderate number of data points
needs to be corrected.

\paragraph{Non-convex loss functions}
Our approach can only guarantee certified unlearning for strongly
convex loss functions that have Lipschitz-continuous gradients. While
both update steps of our approach work well for neural networks with
non-convex functions, as we demonstrate in the empirical evaluation,
they require an additional measure to validate unlearning success.
Fortunately, such external measures are often available, as they
typically provide the basis for characterizing data leakage prior to
its removal.  
%
Similarly, we use the fidelity to measure how
our approach corrects a poisoning attack. 


\paragraph{Unlearning requires detection} 
Finally, we point out that our method requires knowledge of the data
to be removed. Detecting privacy leaks in learning models is a hard
problem outside of the scope of this work. The nature of privacy leaks
depends on the considered data, learning models, and application. For
example, the analysis of \citet{CarLiuErl+19,CarTraWal+21} focuses on
sequential data in generative learning models and cannot be easily
transferred to other learning models or image data. As a result, we
limit this work to repairing leaks rather than finding them.
\section{Conclusion}
\label{sec:conclusions}

Instance-based unlearning is concerned with removing data points from
a learning model \emph{after} training---a task that becomes essential
when users demand the ``right to be forgotten''. However, sensitive
information is often spread across instances, impacting larger
portions of the training data.  Instance-based unlearning is limited
in this setting.  As a remedy, we propose a novel framework for
unlearning features and labels based on the concept of influence
functions. Our approach captures the changes to a learning model in a
closed-form update, providing significant speed-ups over other
approaches.


We demonstrate the effectivity of our approach in a theoretical and
empirical analysis. Based on the concept of differential privacy, we
prove that our framework enables certified unlearning on models with a
strongly convex loss function and evaluate the benefits of our
unlearning strategy in three practical scenarios. In particular, for
generative language models, we are able to remove unintended
memorization while preserving the functionality of the models.

We hope that this work fosters further research on machine unlearning
and sharpens theoretical bounds on privacy in machine
learning. To support this development, we make all our implementations
and datasets publicly available at
{\small\url{https://github.com/alewarne/MachineUnlearning}}.

\section*{Acknowledgment}
We thank Önder Askin for proofreading and valuable comments. This work
was funded by the German Federal Ministry of Education and Research~(BMBF)
under the projects BIFOLD (FKZ01IS18025B) and DataChainSec (FKZ16KIS1700).
Furthermore, the authors acknowledge funding by the Deutsche
Forschungsgemeinschaft (DFG, German Research Foundation) under Germany’s
Excellence Strategy EXC 2092 CASA-390781972 and by the Helmholtz Association
(HGF) within topic ``46.23 Engineering Secure Systems''.

	\bibliography{bib/other,%
		bib/sectubs/sec,%
		bib/sectubs/learn,%
		bib/sectubs/sectubs}

\appendix

\section{Appendix}

\subsection{Stochastic Analysis of Sharding}
\label{appdx:sharding}

To better understand the need for closed-form updates on model
parameters, we examine current sharding strategies and investigate the
circumstances under which they reach their limits.
\citet{BourChaCho+21} propose an unlearning method with the core idea
to train separate models on distinct parts of training data.  While
the authors discuss limitations of their approach like performance
degradation, we perform a stochastic analysis to find upper bounds for
the number of affected data points at which sharding becomes as
efficient as retraining.

In this context, we consider $n$ data instances to unlearn which are
uniformly distributed across $s$ shards. Let $p(n)$ denote the
probability that all shards contain at least one of these samples which
leads to the worst-case scenario of having to retrain all
shards. Since calculating $p(n)$ as stated above is difficult, we
reformulate the task to solve an equivalent problem: We seek the
probability $\hat{p}_k(n)$ that at most $k$ shards remain unaffected
by any sample. We set $k=s$ such that $\hat{p}_s(n)$ indicates the
probability that any combination of $i \in \lbrace 1,\dots,s \rbrace$
shards are unaffected. If this probability is zero there are no
unaffected shards. Hence, this corresponds to the inverse of our
target probability $p(n) = 1 - \hat{p}_s(n)$.

To calculate $\hat{p}_s(n)$, we first determine the probability of
exactly $i$ shards to remain unaffected. In general, there are
$(s-i)^n$ combinations to distribute $n$ samples on the dataset
excluding $i$ shards. Since there are $\binom{s}{i}$ possible ways to
select the $i$ shards to be left out, the total number of combinations
is given by $\binom{s}{i} (s-i)^n$.
However, we cannot simply sum these terms up for different values of $i$
since the unaffected shards in the combinations partly overlap. 
To account for this, we apply the inclusion-exclusion principle
and finally divide the adjusted term by the number of combinations
including all shards:
\begin{equation*}
	\hat{p}_s(n) = \frac{\sum_{i=1}^{s} (-1)^{i+1} \binom{s}{j} (s-i)^n}{s^n}
\end{equation*}

\fig{fig:sharding} in Section~\ref{sec:relatedwork} shows that $p(n)$
quickly reaches one even for low numbers of affected samples. Since
the probability only depends on the number of shards and samples to
unlearn and not on the size of the dataset, we can conclude that
sharding is inefficient when there are many unlearning requests. This
essentially motivates our approach using closed-form updates on model
parameters.

\subsection{Deriving the Update Steps}

In this section, we derive the first-order and second-order update
strategies used in our approach. For a deeper theoretical discussion
of the employed techniques, we recommend the reader the book
of~\citet{BoyVan}.

\subsubsection{First-order update}\label{sec:derivation-gradient}
To derive the first-order update, let us recall the optimization
problem for the corrected learning model from
Section~\ref{sec:approach}:
\begin{align}
	\label{eq:basic1}
	\optmodel_{\epsilon, \unlearn{\point}{\pert}}
	& =
	\argmin_{\model} \bigloss(\model; D) 
	+ \epsilon \loss(\pert, \model)
	- \epsilon \loss(\point, \model) \\
	& = 
	\argmin_{\model} \nonumber
	\bigloss_{\epsilon}(\model; D),
\end{align}
where $\bigloss_{\epsilon}(\model; D)$ is a combined loss function
containing our update and the regularized loss $\bigloss(\model; D)$.
If $\epsilon$ is small and $\loss$ is differentiable with respect to
$\model$, we can approximate $\bigloss_{\epsilon}(\model; D)$ using a
first-order Taylor series at $\optmodel$ by
\begin{align}
	\bigloss_{\epsilon}(\optmodel_{\epsilon, \unlearn{\point}{\pert}}; D)
	\approx \,  &\bigloss(\optmodel; D) \\
	+ &\epsilon  \big(\loss(\pert, \optmodel) - \loss(\point, \optmodel)\big) \nonumber\\
	+ \nonumber &\Delta(\points, \perts) \cdot \Big( \nabla_{\model} \bigloss(\optmodel; D) \\
	& \quad+ \epsilon\big(\gradlossPert - \gradlossPoint\big)\Big). \nonumber 
\end{align}
Since the corrected model
$\optmodel_{\epsilon, \unlearn{\point}{\pert}}$ is a minimum of
$\bigloss_{\epsilon}(\cdot; D)$, we can assume that
\mbox{$\bigloss_{\epsilon}(\optmodel_{\epsilon,
    \unlearn{\point}{\pert}}; D) < \bigloss_{\epsilon}(\optmodel;
  D)$}.
Incorporating this assumption in the Taylor series approximation and
using the condition that $\nabla_{\model} \bigloss(\optmodel; D)=0$,
we now arrive at
\begin{equation}
	\epsilon\Delta(\points,\perts)\cdot\Big(
	\gradlossPert-\gradlossPoint\Big) < 0. \nonumber
\end{equation}

As we have $\epsilon>0$, we can continue to focus on the dot product
of the equation. For two vectors $u,v$ the dot product can be written
as $u\cdot v = \Vert u\Vert\Vert v \Vert \cos(u,v)$ where $\cos(u,v)$
is the cosine between the vectors $u$ and $v$. The minimum of the
cosine is $-1$ which is achieved when $u=-v$, hence we have
\[
\Delta(\points, \perts) = -\big(\gradlossPert-\gradlossPoint\big)
\]

This result indicates that $\gradlossPert-\gradlossPoint$ is the
optimal direction to move starting from $\optmodel$. The actual step
size, however, is unknown and must be adjusted by a small constant
$\tau$ yielding the update step defined in \sect{sec:first-order}:
$$\optmodel_{\epsilon, \unlearn{\point}{\pert}} = \optmodel-\tau
\big(\gradlossPert-\gradlossPoint\big).$$
Due to the linearity of the gradient in this step, the derivation is
equal when multiple points are affected.

\subsubsection{Second-order update}\label{sec:derivation-influence}
If we assume that the loss $\bigloss(\model;D)$ is twice
differentiable and strictly convex, there exists an inverse Hessian
matrix $H_{\optmodel}^{-1}$ and we can proceed to approximate changes
to the learning model using the technique of \citet{CooWei82}.  In
particular, we can determine the optimality conditions for
\cref{eq:basic1}  directly by 
\begin{equation*}
	0 = \nabla\bigloss(\optmodel_{\epsilon, \unlearn{\point}{\pert}}; D) 
	+ \epsilon \nabla 
	\loss(\pert, \optmodel_{\epsilon, \unlearn{\point}{\pert}})
	- \epsilon \nabla 
	\loss(\point, \optmodel_{\epsilon, \unlearn{\point}{\pert}}).
\end{equation*}

If $\epsilon$ is sufficiently small, we can again approximate the
conditions using a first-order Taylor series at $\optmodel$ and obtain
\begin{align*}
	0 \, \approx \, 
	& \nabla \bigloss(\optmodel, D) 
	+ \epsilon \gradlossPert - \epsilon \gradlossPoint 
	\\
	& + (\optmodel_{\epsilon, \unlearn{\point}{\pert}}-\optmodel)  \cdot  
	\nabla^2 \bigloss (\optmodel, D)  \\
	& + \epsilon \nabla^2 \loss(\pert, \optmodel)
	- \epsilon \nabla^2 \loss(\point, \optmodel).
\end{align*}

Since we know that $\nabla\bigloss(\optmodel; D)=0$ by the optimality
of~$\optmodel$, we can rearrange this solution using the Hessian of
the loss function, so that we get 
\begin{align}
	\optmodel_{\epsilon, \unlearn{\point}{\pert}}-\optmodel 
	= -H_{\optmodel}^{-1}\big(\gradlossPert - \gradlossPoint\big)\epsilon,
	\label{eq:deriv_inf}
\end{align}
where we additionally drop all terms in $\mathcal{O}(\epsilon)$. By
expressing this solution in terms of the influence of $\epsilon$ on
the model, we can further simplify it and arrive at
\begin{equation*}
	\frac{\partial\optmodel_{\epsilon, \unlearn{\point}{\pert}}}%
	{\partial\epsilon} \Big \lvert_{\epsilon=0} = 
	-H_{\optmodel}^{-1}\big(\nabla_{\model} \loss(\pert, \optmodel) 
	- \nabla_{\model} \loss(\point, \optmodel)\big).
\end{equation*}
Finally, when using $\epsilon=1$ in \cref{eq:basic1}, the data point
$\point$ is replaced by $\pert$ completely. In this case,
\cref{eq:deriv_inf} directly leads to the second-order update defined
in \sect{sec:second-order-update}
\begin{equation*}
	\optmodel_{\unlearn{\point}{\pert}} 
	\approx 
	\optmodel - H_{\optmodel}^{-1}
	\big(
	\gradlossPert - \gradlossPoint\big).
\end{equation*}

\subsection{Calculating Updates Efficiently}
\label{appdx:algorithm}

To apply second-order updates in practice, we have to avoid storing
the Hessian matrix $H$ explicitly and still be able to compute
$H^{-1}v$. To this end, we rely on the scheme proposed by
\citet{AgaBulHaz17} for computing expressions of the form
$H^{-1}v$. This scheme requires to only calculate $Hv$ and avoids
storing $H^{-1}$. The resulting \emph{Hessian-Vector-Products} (HVPs)
allow us to calculate $Hv$ efficiently by making use of the linearity
of the gradient
\begin{equation*}
	Hv=\nabla_{\theta}^2 \bigloss(\theta^*;D)v=\nabla_{\theta}\big(\nabla_{\theta}\bigloss(\theta^*;D) v\big).
\end{equation*}
If we denote first $j$ terms of the Taylor expansion of $H^{-1}$ by
$H_j^{-1} = \sum_{i=0}^j (I-H)^i$, we can recursively define the
approximation $H_j^{-1} = I + (I-H)H_{j-1}^{-1}$. Now, if
$\vert \lambda_i\vert <1$ for all eigenvalues $\lambda_i$ of $H$, we
have $H_j^{-1}\rightarrow H^{-1}$ for $j\rightarrow \infty$. To ensure
this convergence, we add a small damping term $\lambda$ to the
diagonal of $H$ and scale down the loss function by some constant
which does not change the optimal parameters $\theta^*$.
We can then formulate the following algorithm for computing an
approximation of $H^{-1}v$: Given data points $z_1,\dots,z_t$ sampled
from $D$, we define the iterative updates
\begin{align*}
	\tilde{H}_0^{-1}v &= v,\\
	\tilde{H}_j^{-1}v &= v + \big(I-\nabla_{\theta}^2L(z_i,\theta^*)\big)\tilde{H}_{j-1}^{-1}v.
\end{align*}
In each update step, $H$ is estimated using a single data point and we
can use HVPs to evaluate
$\nabla_{\theta}^2L(z_i,\theta^*)\tilde{H}_{j-1}^{-1}v$ efficiently in
$\mathcal{O}(p)$ as demonstrated by \citet{Pea94}. 

Averaging batches of data points further speeds up the
approximation. Choosing $t$ large enough so that the updates converge
and averaging $r$ runs to reduce the variance of the results, we
obtain $\tilde{H}_t^{-1}v$ as our final estimate of $H^{-1}v$ in
$\mathcal{O}(rtp)$ of time. The pseudo-code in Algorithm 1 summarizes
how we compute the second-order update.

\begin{algorithm}[t]
	\label{algo:paramupdate}
	\DontPrintSemicolon
	
	\KwInput{model $\model^*$, loss functions $\bigloss$ and $\loss$,
		order o, unlearning rate $\tau$, batch-size B, iterations m,
		damping d, scale s, repetitions r}
	\KwOutput{Parameter update $\Delta(Z,\tilde{Z})$}
	\KwData{$D$, $D'$, $Z$, $\tilde{Z}$}
	$g_1 = \sum_{\pert \in \perts}\gradlossPert$, 
	$g_2 = \sum_{\point \in \points} \gradlossPoint$\\
	$v=g_1-g_2$\\
	\If{o == 1}
	{
		
		$\Delta = -\tau v$
	}
	\Else
	{
		$\Delta = 0$\\
		
		\For{i=1:r}
		{
			$\model_{\text{new}} = 0$\\
			\For{j=1:m}    
			{ 
				batch = sample(D', size=B)\\
				hvp =$\nabla_{\theta}\big(v^T\nabla_{\theta}
				L(\text{batch}, \model^*)\big)$
				$\model_{\text{new}} = v + (1-d) \model_{\text{new}}
				-\text{hvp}/s$
			}
			$\Delta = \Delta + \model_{\text{new}} / r$
	}}
	return $\Delta$
	\caption{Parameter update}
\end{algorithm}

\subsection{Proofs for Certified Unlearning}
\label{sec:proofs}

We continue to present the proofs for certified unlearning of our
approach and, in particular, the bounds of the gradient residual used
in \cref{sec:cert-unlearning}.
First, let us recall \cref{thm:thm1} from \cref{sec:cert-unle-feat}.

\newtheorem{thm-recall}{Theorem}
\begin{thm-recall}
	Assume that $\Vert x_i\Vert_2 \leq 1$ for all data points and the
gradient
$\nabla\loss(z,\model)$ is $\gamma_z$-Lipschitz with respect to $z$ at
$\optmodel$ and \mbox{$\gamma$-Lipschitz} with respect to
$\model$. Further let $\perts$ change the features $j,\dots,j+F$ by
magnitudes at most $m_j,\dots,m_{j+F}$. If $M=\sum_{j=1}^{F} m_j$ the
following upper bounds hold:
	
\begin{enumerate}
	\item For the first-order update of our approach, we have
	    $$\big\Vert\nabla\bigloss\big(\optmodel_
		{\points\rightarrow\perts}, D^\prime\big)\big\Vert_2 \leq
		(1+\tau\gamma n)\gamma_zM\vert\points\vert$$
	\item If $\nabla^2\loss(z,\model)$ is $\gamma''$-Lipschitz with
	respect to $\model$, we have 
		$$\big\Vert\nabla\bigloss\big(\optmodel_
		{\points\rightarrow\perts}, D^\prime\big)\big\Vert_2 \leq \gamma''\Big(\frac{M\gamma_z}{\lambda}\Big)^2 n\vert\points\vert^2$$
		for the second-order update of our approach.

\end{enumerate}

\end{thm-recall}

To prove this theorem, we begin by introducing a small lemma which is
useful for investigating the gradient residual of the optimal learning
model $\optmodel$ on a dataset $D^\prime$.
\begin{lem}
	\label{lem:appdx-lazy_bounds}
	Given a radius $R > 0$ with $\Vert \delta_i\Vert_2 \leq R$, a gradient
$\nabla\loss(z,\model)$ that is $\gamma_z$-Lipschitz with respect to
$z$, and a learning model $\model^*$, we have
$$\big\Vert \nabla\bigloss\big(\optmodel,D^\prime \big)\big\Vert_2\leq
		R\gamma_z\vert\points\vert.$$
\end{lem}

\begin{proof}	
	By definition, we have
	\begin{align}
		\nabla\bigloss(\theta^*; D^\prime) &= 
		\sum_{z\in D^\prime}\nabla \loss\big(z,\optmodel\big)
		+ \lambda\optmodel.\nonumber
	\end{align}
	We can now split the dataset $D^\prime$ into the set of affected data points~$\perts$ and the remaining data as follows
	\begin{align}
		\nabla\bigloss(\theta^*; D^\prime)  
		&= \sum_{\point\in D^\prime\setminus \perts}
		\nabla	\loss\big(\point,\optmodel\big)+
		\sum_{\pert\in \perts}\nabla \loss\big(\pert,\optmodel\big)
		+ \lambda\optmodel\nonumber\\
		&= \sum_{\point\in D\setminus \points}
		\nabla	\loss\big(z,\optmodel\big)+
		\sum_{\pert\in \perts}\nabla \loss\big(\pert,\optmodel\big)
		+ \lambda\optmodel.\nonumber
	\end{align}	
	By applying a zero addition and leveraging the optimality of
	$\model^*$ on $D$, we then express the
	gradient as follows
	\begin{align}	
		\nabla\bigloss(\theta^*; D^\prime)  
		&=0 + \sum_{\point_i\in \points} 
		\nabla\loss\big(\point_i+\delta_i, \optmodel\big) -
		\nabla \loss\big(\point_i, \optmodel\big).\label{eqn:lazy}
	\end{align}
	Finally, using the \lipschitz of $\nabla\loss$,
	we get 
	%
	\begin{align}
		\big\Vert \nabla\bigloss\big(\optmodel, D^\prime\big)\big\Vert_2
		&\leq\sum_{\point_i\in \points}\big\Vert\nabla\loss\big(\point_i+\delta_i, \optmodel\big) -
		\nabla \loss\big(\point_i, \optmodel\big)\big\Vert_2\nonumber\\
		&\leq\sum_{x_i,y_i\in \points}\gamma_z\big\Vert 
		\delta_i\big\Vert_2\nonumber\leq M\gamma_z \vert\points\vert.\nonumber
	\end{align}
\end{proof}

We proceed to prove the update bounds of \cref{thm:thm1}. The proof is
structured in two parts, where we start with investigating the first
case and then proceed with the second case of the theorem.


\begin{proof}~(Case 1)
	For the first-order update, we recall that
	$$
	\theta^*_{\points\rightarrow\perts} \, = \theta^*
	-\tau G(\points, \perts)
	$$ where $\tau\geq 0$ is the unlearning rate and we have
	$$
	G(\points, \perts) \, =\sum_{z_i \in\points}
	\nabla\loss(z_i + \delta_i, \model) -
	\nabla\loss(z_i, \model)
	$$  
	Consequently, we seek to bound the norm
	of
	\begin{equation*}
		\nabla \bigloss\big(\theta^*_{\unlearn{\points}{\perts}},
		D^\prime\big) = \nabla \bigloss\big(\theta^*-\tau
		G(\points, \perts),D^\prime\big).
	\end{equation*}
	
	\noindent By Taylor's theorem, there exists a constant
	$\eta\in [0,1]$ and a parameter
	$\theta^*_{\eta}=\theta^*-\eta \tau G(\points, \perts)$ such
	that
	\begin{align*}
		\nabla \bigloss\big(\optmodel_{\unlearn{\points}{\perts}}
		,D^\prime\big)
		&=\nabla\bigloss\big(\optmodel,D^\prime\big) \\
		&+\nabla^2\bigloss\big(
		\optmodel+\eta(\optmodel_{\unlearn{\points}{\perts}}-\optmodel),
		D^\prime\big) \big(\optmodel_{\unlearn{\points}{\perts}}-
		\optmodel\big)\\
		&=\nabla\bigloss\big(
		\optmodel,D^\prime\big) - \tau H_{\model^*_{\eta}}
		G(\points,\perts).
	\end{align*}
	In the proof of \cref{lem:appdx-lazy_bounds} we show that
	$
	\nabla\bigloss\big(\optmodel,D^\prime\big) = G(\points,\perts)
	$	
	and thus we get
	\begin{align*}
		\big\Vert\nabla\bigloss\big(\optmodel_{\unlearn{\points}{\perts}},
		D^\prime\big)\big\Vert_2 &= \Vert G(\points,\perts) - \tau
		H_{\model^*_{\eta}}	G(\points,\perts)\Vert_2\\
		&= \big\Vert\big(I-\tau H_{\optmodel_{\eta}}\big)
		G(\points, \perts)\big\Vert_2\\
		&\leq \big\Vert I-\tau H_{\optmodel_{\eta}}\big\Vert_2
		\big\Vert G(\points, \perts)\big\Vert_2.
	\end{align*}
	
	\noindent Due to the $\gamma$-\lipschitz of the gradient $\nabla\loss$, we
	have $\Vert H_{\optmodel_{\eta}}\Vert_2 \leq n\gamma$ and thus
	\begin{align*}
		\Vert I-\tau H_{\optmodel_{\eta}}\Vert_2 &\leq 1+\tau \gamma n
	\end{align*}
	which, with the help of \cref{lem:appdx-lazy_bounds}, yields
	the final bound for the first-order update	
	$$\big\Vert\nabla\bigloss\big(\optmodel_
	{\points\rightarrow\perts}, D^\prime\big)\big\Vert_2 \leq
	(1+\tau\gamma n)M\gamma_z\vert\points\vert.$$
\end{proof}

\begin{proof}~(Case 2)
         For the second-order update of our approach, we recall that
	$$\theta^*_	{\points\rightarrow\perts} = \theta^*
	-H^{-1}_{\theta^*}G(\points, \perts).$$
	Similar to the proof for the first-order update, there exists some
	$\eta\in [0,1]$ and a parameter
	$\theta^*_{\eta}=\theta^*-\eta H^{-1}_{\theta^*}G(\points, \perts)$
	such that	
	\begin{align*}
		\nabla \bigloss\big(\optmodel_{ \unlearn{\points}{\perts}},D^\prime\big)
		&=\nabla\bigloss\big(
		\optmodel,D^\prime\big) \\
		&+ \nabla^2\bigloss\big(
		\optmodel+\eta(\optmodel_{\unlearn{\points}{\perts}}-\optmodel),D^\prime\big)\big(\optmodel_{\unlearn{\points}{\perts}}-\optmodel\big)\\
		&=\nabla\bigloss\big(
		\optmodel,D^\prime\big) - H_{\model^*_{\eta}}H^{-1}_{\optmodel}
		G(\points, \perts).
	\end{align*}
	\noindent Using again that	
	$\nabla\bigloss\big(\optmodel,D^\prime\big) = G(\points,\perts)$
	we arrive at	
	\begin{align*}
		\big\Vert\nabla\bigloss\big(\optmodel_{\unlearn{\points}{\perts}},
		D^\prime\big)\big\Vert_2 &= \big\Vert G(\points, \perts) -
		H_{\optmodel_{\eta}}H^{-1}_{\optmodel}G(\points, \perts)
		\big\Vert_2\\
		&= \big\Vert\big(H_{\optmodel}-H_{\optmodel_{\eta}}\big)
		H^{-1}_{\optmodel} G(\points, \perts)\big\Vert_2\\
		&\leq \big\Vert H_{\theta^*}-H_{\optmodel_{\eta}}\big\Vert_2
		\big\Vert H^{-1}_{\optmodel}\big\Vert_2\big\Vert G(\points,
		\perts)\big\Vert_2
	\end{align*}
	
	\noindent The $\lambda$-strong convexity of $\bigloss$ ensures that
	$\Vert H^{-1}_{\theta^*}\Vert_2\leq \frac{1}{\lambda}$. In
	addition to
	$\Vert G(\points,\perts) \Vert_2 \leq M\gamma_z \vert\points\vert$,
	it remains to bound the difference between the Hessians.  Using the
	\lipschitz of the gradient $\nabla^2\loss$ for 
	$z\in D^\prime$, we first get
	\begin{align*}
		\big\Vert \nabla^2\loss\big(z,\optmodel) - 
		\nabla^2\loss\big(z, \optmodel_\eta\big)\big\Vert_2
		&\leq \gamma'' \big\Vert \optmodel -
		\model^*_\eta\big\Vert_2\\
		&\leq \gamma''\big\Vert H^{-1}_{\optmodel}\big\Vert_2
		\big\Vert G(\points,\perts)\big\Vert_2
	\end{align*}
	and then for the Hessians obtain 
	\begin{align*}
		\big\Vert H_{\theta^*}-H_{\theta^*_{\eta}}\big\Vert_2
		&= \sum_{z\in D}\big\Vert
		\nabla^2\loss\big(z,\optmodel) -
		\nabla^2\loss\big(z,\theta^*_\eta\big)
		\big\Vert_2\\
		&\leq n \gamma'' \big\Vert
		H^{-1}_{\theta^*}\big\Vert_2 \big\Vert G(\points,\perts)\big\Vert_2.
	\end{align*}
	Combining all results finally yields the theoretical bound for the
	second-order update of our approach
	\begin{align*}
		\big\Vert\nabla\bigloss\big(\optmodel_{\unlearn{\points}{\perts}},
		D^\prime\big)\big\Vert_2 &\leq \big\Vert
		H_{\theta^*}-H_{\theta^*_{\eta}}\big\Vert_2
		\big\Vert H^{-1}_{\theta^*}\big\Vert_2\big\Vert
		G(\points,\perts)\big\Vert_2\\
		&\leq n \gamma''
		\big\Vert H^{-1}_{\optmodel}\big\Vert_2^2
		\big\Vert G(\points, \perts)\big\Vert_2^2\\
		&\leq \gamma''\Big(\frac{M\gamma_z}{\lambda}\Big)^2 n\vert\points\vert^2
	\end{align*}
\end{proof}

\begin{thm}
	Let $\mathcal{A}$ be the learning algorithm that returns the
unique minimum of $\bigloss_b(\theta;D^\prime)$ and let
$\mathcal{U}$ be an unlearning method that produces a model
$\model_{\mathcal{U}}$. If
$\Vert \nabla\bigloss(\theta_\mathcal{U};D^\prime)\Vert_2 \leq
\epsilon'$
for some $\epsilon' >0$ we have the following guarantees.
\begin{enumerate}
	\item If $b$ is drawn from a distribution with density
	$p(b)=e^{-\frac{\epsilon}{\epsilon'}\Vert b\Vert_2}$ then
	$\mathcal{U}$ performs $\epsilon$-certified unlearning for
	$\mathcal{A}$.
	
	\item If $p\sim \mathcal{N}(0, c\epsilon'/\epsilon)^d$ for
	some $c>0$ then $\mathcal{U}$ performs $(\epsilon,
	\delta)$-certified unlearning for $\mathcal{A}$ with
	$\delta=1.5e^{-c^2/2}$.
\end{enumerate}
\end{thm}

\begin{proof}
	The proofs work similarly to the sensitivity proofs for differential
	privacy as presented in~\citet{DwoRot14}.
	\begin{enumerate}
        \setlength{\itemsep}{4pt}
		\item Given $b_1$ and $b_2$ with $\Vert b_1-b_2\Vert_2 \leq
		\epsilon'$. By the construction of the density $p$, we have
		\begin{align*}
			\frac{p(b_1)}{p(b_2)} = e^{-\frac{\epsilon}{\epsilon'}\big(\Vert b_1\Vert_2 - \Vert b_2\Vert_2\big)}\leq e^{\frac{\epsilon}{\epsilon'}\big(\Vert b_1 - b_2\Vert_2\big)}\leq e^\epsilon.
		\end{align*}
                If we now apply \cref{thm:thm2}, we can finalize the
                proof for the first case.

		\item The second proof is similar to Theorem 3.22 in~\citet{DwoRot14} using
		$\Delta_2(f)=\epsilon'$ which yields that with probability at least
		$1-\delta$ we have $e^{-\epsilon}\leq \frac{p(b_1)}{p(b_2)}\leq
		e^\epsilon$. Applying \cref{thm:thm2} afterwards again finalizes the
		proof.
	\end{enumerate}
\end{proof}

\subsection{Relation to Differential Privacy}
\label{sect:relation-dp}

Our definition of certified unlearning shares interesting similarities
with the concept of differential privacy~\citep{Dwo06} that we
highlight in the following. First, let us recall the definition of
differential privacy for a learning algorithm $\mathcal{A}$:

\begin{defin}
	\label{def-dp}
	Given some $\epsilon >0$, a learning algorithm $\mathcal{A}$
	is said to be $\epsilon$-differentially private ($\epsilon$-DP) if
	\begin{equation*}
		\label{eq:def-dp}
		e^{-\epsilon} \leq \frac{P\big(\mathcal{A}(D)\in\mathcal{T}\big)}{P\big(\mathcal{A}(D')\in\mathcal{T}\big)}\leq e^{\epsilon}
	\end{equation*}
	holds for all $\mathcal{T} \subset \Model$ and datasets $D$,
        $D'$ that differ in one sample.  That is, we consider
        $\vert D\vert=\vert D'\vert$ where one sample has been
        replaced. This is denoted as ``bounded differential privacy''
        in literature~\citep{DesPej20, KifMac11}.
\end{defin}

By this definition, differential privacy is a \emph{sufficient}
condition for certified unlearning. We obtain the same bound by simply
setting the unlearning method $\mathcal{U}$ to the identity function
in \cref{def-ecr}. That is, if we cannot distinguish whether
$\mathcal{A}$ was trained with a point $z$ or its modification
$z_\delta$, we do not need to worry about unlearning $z$ later. This
result is also obtained by \cref{thm:thm1} when we set the unlearning
rate $\tau$ to zero and no model updates are performed during
unlearning.

\begin{table*}[t]
	\begin{center}
	\caption{Average runtime when removing \num{100} random
		combinations of features for the different datasets.}
	\label{tab:appdx-runtime}
	\tablesize \setlength{\tabcolsep}{4pt}
	\begin{tabular}{
			l
			S[table-format = 2.1e1]
			S[table-format = 1.2]
			S[table-format = 3.1]
			l
			S[table-format = 2.1e1]
			S[table-format = 1.2]
			S[table-format = 2.1]
			l
			S[table-format = 2.1e1]
			S[table-format = 1.2]
			S[table-format = 3.1]
		}
		\toprule
		& \multicolumn{3}{c}{\bfseries \enron} & & \multicolumn{3}{c}{\bfseries \diabetis} & &
		\multicolumn{3}{c}{\bfseries \adult}\\
		\toprule
		\bfseries Method & {\bfseries Gradients} \hspace{1mm} & {\bfseries Runtime} \hspace{1mm} & {\bfseries Speed-up} & &{\bfseries Gradients} \hspace{1mm} & {\bfseries Runtime} \hspace{1mm} & {\bfseries Speed-up}&& {\bfseries Gradients} \hspace{1mm} & {\bfseries Runtime} \hspace{1mm} & {\bfseries Speed-up}\\
		\midrule
		\retraining & 1.4e6 & 1.52 s & {~~---}& & 1.1e4 & 6.71 ms & {~~---} & &5.1e6 & 2.94 s & {~~---} \\ 
		\sharding{5} & 4.1e5 & 0.56 s & 2.7\si{\speedup}&& 9.8e3 & 29.55 ms & 0.2 \si{\speedup} && 2.4e6 & 1.65 s & 1.8 \si{\speedup}\\
		\finetuning & 2.6e4 & 0.80 s & 1.9\si{\speedup} && 6.1e2 & 0.60 ms & 11.2 \si{\speedup}&& 3.9e4 & 0.09 s & 32.7 \si{\speedup}\\
		\midrule
		\firstorder  & 1.1e4 & 0.04 ms & 38.0 \si{\speedup}&& 1.0e2 & 0.10 ms & 67.1 \si{\speedup}&& 1.0e2 & 4.87 ms & 603.7 \si{\speedup}\\
		\secondorder  & 6.5e4 & 5.42 s & 0.3 \si{\speedup}&& 1.3e4 & 0.23 ms & 29.2 \si{\speedup}&& 7.8e4 & 0.03 s & 98.0 \si{\speedup} \\
		\bottomrule
	\end{tabular}
\end{center}
\end{table*}



Consequently, certified unlearning can be obtained through DP, yet the
learning model's performance may suffer when enforcing strong privacy
guarantees~\citep[see][]{ChaMonSar11, AbaChuGoo+16}. In
\sect{sec:empirical-analysis}, we demonstrate that the models obtained
using our update strategies are much closer to $\mathcal{A}(D')$ in
terms of performance compared to DP alone. In this light, our approach
to certified unlearning can be seen as a compromise between the high
privacy guarantees of DP and the optimal performance achievable
through costly re-training.

\subsection{Multiple Unlearning Steps}
\label{sect:multiple-steps}

So far, we have considered unlearning as a one-shot strategy. That is,
all changes are incorporated in $\perts$ before performing an
update. However, \cref{thm:thm1} shows that the error of the updates
rises linearly with the number of affected points and the size of the
total perturbation. Instead of performing a single update, it thus
becomes possible to split $\perts$ into $T$ subsets and conduct $T$
consecutive updates. In terms of run-time it is easy to see that the
total number of gradient computations remains the same for the
first-order update. For the second-order strategy, however, multiple
updates require calculating the inverse Hessian for each intermediate
step, which increases the computational effort.  An empirical
comparison of our approach with multiple update steps is presented in
\sect{appdx-sequential}.


In terms of unlearning certifications, we can extend \cref{thm:thm1}
and show that the gradient residual bound after $T$ update steps
remains smaller than $TC$, where $C$ is the bound of a single step: If
$\model_t$ is the $t$-th solution obtained by our updates with
gradient residual $r_t$ then $\model_t$ is an exact solution of the
loss function $\bigloss_b(\model, D')-r_t^T\model$ by
construction. This allows applying \cref{thm:thm1} to each $\model_t$
and obtain the bound $TC$ by the triangle inequality. Therefore, the
gradient residual bound rises linearly in the number of applications
of $\mathcal{U}$. This result is in line with the composition theorem
of differential privacy research~\citep{DwoMcsKob+06, DwoRotVad10,
	DwoLei09} which states that applying an $\epsilon$-DP algorithm $n$
times results in a $n\epsilon$-DP algorithm.

\subsection{Evaluation of Certified Unlearning}
\label{appdx:LinearEval}

\subsubsection{Fidelity evaluation with loss}
\label{appdx:fidelity-evaluation}
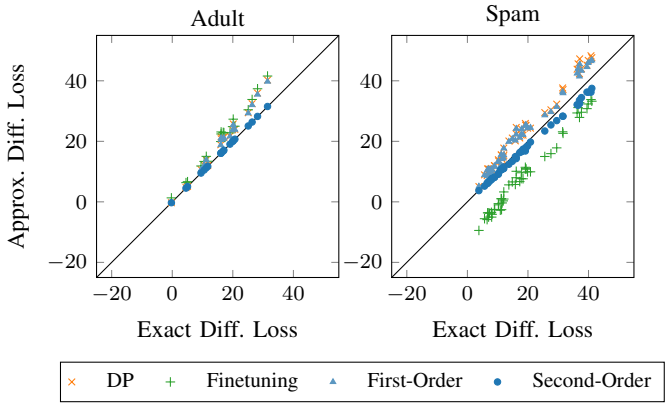
\begin{figure}[!h]
	\definecolor{DP}{HTML}{ff7f0e}
\definecolor{Finetuning}{HTML}{2ca02c}
\definecolor{Firstorder}{HTML}{5B97C1}
\definecolor{Secondorder}{HTML}{1f77b4}

\begin{tikzpicture}
    \pgfplotsset{footnotesize,samples=10}
    \begin{groupplot}[group style = {group size = 2 by 1, horizontal sep = 20pt, vertical sep = 15pt}, width = 0.54\linewidth, height = 0.54\linewidth]
        \nextgroupplot[
            title={Adult},
            ylabel={Approx. Diff. Loss},
            ymin=-25, ymax=55,
            xmin=-25, xmax=55,
            ytick={-20,0,20,40},
            xtick={-20,0,20,40},
            xlabel=Exact Diff. Loss
            ]
            \addplot[only marks, color=DP, mark=x, mark size=1.8pt] table [x=Retraining, y=DP, col sep=comma]{figures/Unlearning/diff_test_loss/scatter_data_adult_1000.csv};

            \addplot[only marks, color=Finetuning, mark=+, mark size=1.8pt] table [x=Retraining, y=FineTuning, col sep=comma]{figures/Unlearning/diff_test_loss/scatter_data_adult_1000.csv};

            \addplot[only marks, color=Firstorder, mark=triangle*, mark size=1.2pt] table [x=Retraining, y=FirstOrder, col sep=comma]{figures/Unlearning/diff_test_loss/scatter_data_adult_1000.csv};

            \addplot[only marks, color=Secondorder, mark=*, mark size=1.2pt] table [x=Retraining, y=SecondOrder, col sep=comma]{figures/Unlearning/diff_test_loss/scatter_data_adult_1000.csv};

            \addplot [color=black] coordinates {(-60,-60)(60,60)};

        \nextgroupplot[
            title={Spam},
            xlabel=Exact Diff. Loss,
            ymin=-25, ymax=55,
            xmin=-25, xmax=55,
            ytick={-20,0,20,40},
            xtick={-20,0,20,40},
            legend style = { column sep = 10pt, legend columns = -1, legend to name = grouplegend,},
            ]
            \addplot[only marks, color=DP, mark=x, mark size=1.8pt] table [x=Retraining, y=DP, col sep=comma]{figures/Unlearning/diff_test_loss/scatter_data_enron_100.csv};
            \addlegendentry{DP}%

            \addplot[only marks, color=Finetuning, mark=+, mark size=1.8pt] table [x=Retraining, y=FineTuning, col sep=comma]{figures/Unlearning/diff_test_loss/scatter_data_enron_100.csv};
            \addlegendentry{Finetuning}%

            \addplot[only marks, color=Firstorder, mark=triangle*, mark size=1.2pt] table [x=Retraining, y=FirstOrder, col sep=comma]{figures/Unlearning/diff_test_loss/scatter_data_enron_100.csv};
            \addlegendentry{First-Order}%

            \addplot[only marks, color=Secondorder, mark=*, mark size=1.2pt] table [x=Retraining, y=SecondOrder, col sep=comma]{figures/Unlearning/diff_test_loss/scatter_data_enron_100.csv};
            \addlegendentry{Second-Order}%

            \addplot [color=black] coordinates {(-60,-60)(60,60)};
        
    \end{groupplot}
    \node at ($(group c2r1) + (-2cm ,-3.cm)$) {\ref{grouplegend}}; 
\end{tikzpicture}
	\caption{Difference in test loss between retraining and
		unlearning when removing or changing random combinations of
		features. For perfect unlearning the results would lie on
		the identity line.}
	\label{fig:scatter-loss-appdx}
\end{figure}

In \sect{sec:empirical-analysis}, we evaluate the fidelity of the
retrained model with different approaches using the difference in test
loss presented in a scatter plot. \fig{fig:scatter-loss-appdx} shows
the plots for the \adult and \enron dataset which we omitted
previously due to space limitations. As for the other datasets, it is
apparent that the second-order update has the highest correlation with
the retrained model, i.e., the points are closest to the identity
line.  We also see less variance in terms of deviation from the
identity line compared to the other approaches.

\subsubsection{Efficiency evaluation}
\label{appdx:efficiency}
In Section~\ref{sec:empirical-analysis}, we present the runtime
evaluation only for the \drebin dataset. We show the evaluations for
the remaining datasets in~\cref{tab:appdx-runtime}. The entries are
based on the experiments regarding fidelity where we removed or
replaced \num{100} combinations of \num{100} features from the
datasets. It can be seen that the computation of the inverse Hessian
is faster than retraining in case of the \diabetis and \adult datasets
making the second-order update very efficient in these cases.

\subsubsection{Sequential unlearning steps}
\label{appdx-sequential}
So far, we have presented our approach as one-shot unlearning, that
is, all perturbed samples are collected in the set $\perts$ and the
update is performed on one step. The theoretical analysis in
\sect{sec:cert-unlearning}, however, shows that the error in
approximation and the gradient residual norm rise with the size of
$\perts$. Therefore, a practitioner might want to perform the updates
in smaller portions to keep the error in each update
small. 
%
To evaluate this setting, we repeat the previous experiment but now
split the update into \num{10} steps of equal size.


\begin{figure}[h]
        \vspace{3pt}
	\input{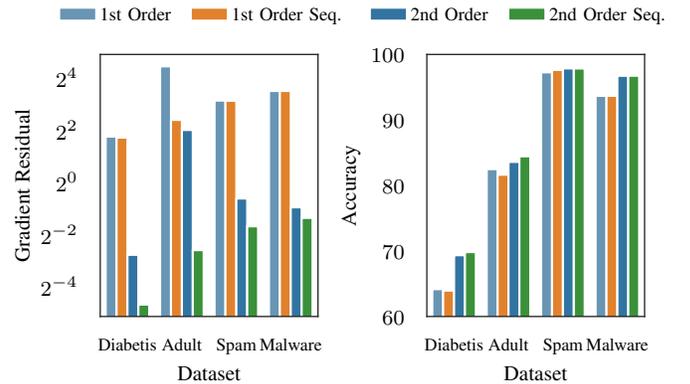}
	\caption{Average accuracy of the corrected models (in \perc{})
		when removing \num{100} features at once or
		sequentially in \num{10} steps.}
	\label{fig:fidelity-multiple-steps}
\end{figure} 

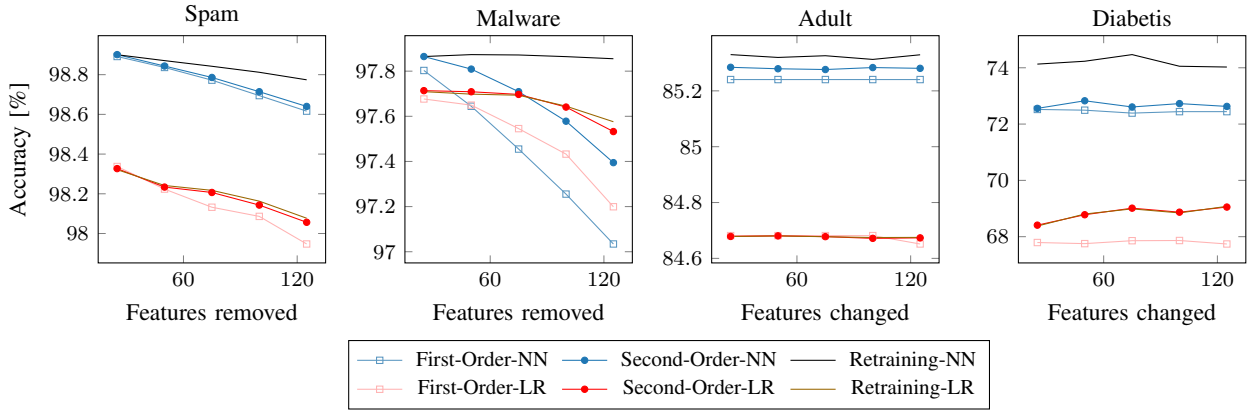
\begin{figure*}[t!]
	\centering
	\subfloat{\definecolor{Finetuning}{HTML}{2ca02c}
\definecolor{Firstorder}{HTML}{5B97C1}
\definecolor{Secondorder}{HTML}{1f77b4}
\definecolor{Retraining}{HTML}{000000}

\definecolor{Retraining-LR}{HTML}{996600}
\definecolor{Firstorder-LR}{HTML}{ffb3b3}
\definecolor{Secondorder-LR}{HTML}{ff0000}

\begin{tikzpicture}
    \pgfplotsset{footnotesize,samples=10}
    \begin{groupplot}[group style = {group size = 4 by 1, horizontal sep = 30pt, vertical sep=20pt}, width = 0.25\linewidth, height = 0.25\linewidth]
        \nextgroupplot[
            title = {Spam},
            legend style = { column sep = 3pt, legend columns = 3, legend to name = grouplegend,},
            xlabel={Features removed},
            ylabel={Accuracy [\%]},
            xtick={60,120},
            legend pos= south west,
            ]

            \addplot[name path=f,color=Firstorder, mark=square, mark size=1.2pt] plot [] table [x=x, y=first-order-mean, col sep=comma]{figures/Unlearning/neural_network_experiments/enron_results.csv};
            \addlegendentry{First-Order-NN}%

            \addplot[name path=f,color=Secondorder, mark=*, mark size=1.2pt] plot [] table [x=x, y=second-order-mean, col sep=comma]{figures/Unlearning/neural_network_experiments/enron_results.csv};
            \addlegendentry{Second-Order-NN}%

            \addplot[name path=f,color=Retraining, mark size=1.2pt] plot [] table [x=x, y=retraining-mean, col sep=comma]{figures/Unlearning/neural_network_experiments/enron_results.csv};
            \addlegendentry{Retraining-NN}%

            \addplot[name path=f,color=Firstorder-LR, mark=square, mark size=1.2pt] plot [] table [x=x, y=1st-Order-mean, col sep=comma]{figures/Unlearning/neural_network_experiments/enron_result_lr.csv};
            \addlegendentry{First-Order-LR}

            \addplot[name path=f,color=Secondorder-LR, mark=*, mark size=1.2pt] plot [] table [x=x, y=2nd-Order-mean, col sep=comma]{figures/Unlearning/neural_network_experiments/enron_result_lr.csv};
            \addlegendentry{Second-Order-LR}%

            \addplot[name path=f,color=Retraining-LR, mark size=1.2pt] plot [] table [x=x, y=Retrained-mean, col sep=comma]{figures/Unlearning/neural_network_experiments/enron_result_lr.csv};
            \addlegendentry{Retraining-LR}

        \nextgroupplot[
            title = {Malware},
            xlabel={Features removed},
            xtick={60,120},
            ]

            \addplot[name path=f,color=Firstorder, mark=square, mark size=1.2pt] plot [] table [x=x, y=first-order-mean, col sep=comma]{figures/Unlearning/neural_network_experiments/drebin_results.csv};

            \addplot[name path=f,color=Secondorder, mark=*, mark size=1.2pt] plot [] table [x=x, y=second-order-mean, col sep=comma]{figures/Unlearning/neural_network_experiments/drebin_results.csv};

            \addplot[name path=f,color=Retraining, mark size=1.2pt] plot [] table [x=x, y=retraining-mean, col sep=comma]{figures/Unlearning/neural_network_experiments/drebin_results.csv};

            \addplot[name path=f,color=Firstorder-LR, mark=square, mark size=1.2pt] plot [] table [x=x, y=1st-Order-mean, col sep=comma]{figures/Unlearning/neural_network_experiments/drebin_result_lr.csv};

            \addplot[name path=f,color=Secondorder-LR, mark=*, mark size=1.2pt] plot [] table [x=x, y=2nd-Order-mean, col sep=comma]{figures/Unlearning/neural_network_experiments/drebin_result_lr.csv};   

            \addplot[name path=f,color=Retraining-LR, mark size=1.2pt] plot [] table [x=x, y=Retrained-mean, col sep=comma]{figures/Unlearning/neural_network_experiments/drebin_result_lr.csv};

        \nextgroupplot[
            title = {Adult},
            xlabel={Features changed},
            xtick={60,120},
            ]

            \addplot[name path=f,color=Firstorder, mark=square, mark size=1.2pt] plot [] table [x=x, y=first-order-mean, col sep=comma]{figures/Unlearning/neural_network_experiments/adult_results.csv};

            \addplot[name path=f,color=Secondorder, mark=*, mark size=1.2pt] plot [] table [x=x, y=second-order-mean, col sep=comma]{figures/Unlearning/neural_network_experiments/adult_results.csv};

            \addplot[name path=f,color=Retraining, mark size=1.2pt] plot [] table [x=x, y=retraining-mean, col sep=comma]{figures/Unlearning/neural_network_experiments/adult_results.csv};

            \addplot[name path=f,color=Firstorder-LR, mark=square, mark size=1.2pt] plot [] table [x=x, y=1st-Order-mean, col sep=comma]{figures/Unlearning/neural_network_experiments/adult_result_lr.csv};

            \addplot[name path=f,color=Secondorder-LR, mark=*, mark size=1.2pt] plot [] table [x=x, y=2nd-Order-mean, col sep=comma]{figures/Unlearning/neural_network_experiments/adult_result_lr.csv};   

            \addplot[name path=f,color=Retraining-LR, mark size=1.2pt] plot [] table [x=x, y=Retrained-mean, col sep=comma]{figures/Unlearning/neural_network_experiments/adult_result_lr.csv};

        \nextgroupplot[
            title = {Diabetis},
            xlabel={Features changed},
            xtick={60,120},
            ]

            \addplot[name path=f,color=Firstorder, mark=square, mark size=1.2pt] plot [] table [x=x, y=first-order-mean, col sep=comma]{figures/Unlearning/neural_network_experiments/diabetis_results.csv};

            \addplot[name path=f,color=Secondorder, mark=*, mark size=1.2pt] plot [] table [x=x, y=second-order-mean, col sep=comma]{figures/Unlearning/neural_network_experiments/diabetis_results.csv};

            \addplot[name path=f,color=Retraining, mark size=1.2pt] plot [] table [x=x, y=retraining-mean, col sep=comma]{figures/Unlearning/neural_network_experiments/diabetis_results.csv};

            \addplot[name path=f,color=Firstorder-LR, mark=square, mark size=1.2pt] plot [] table [x=x, y=1st-Order-mean, col sep=comma]{figures/Unlearning/neural_network_experiments/diabetis_result_lr.csv};

            \addplot[name path=f,color=Secondorder-LR, mark=*, mark size=1.2pt] plot [] table [x=x, y=2nd-Order-mean, col sep=comma]{figures/Unlearning/neural_network_experiments/diabetis_result_lr.csv};   

            \addplot[name path=f,color=Retraining-LR, mark size=1.2pt] plot [] table [x=x, y=Retrained-mean, col sep=comma]{figures/Unlearning/neural_network_experiments/diabetis_result_lr.csv};
        
    \end{groupplot}
    \node at ($(group c2r1) + (2cm ,-3.cm)$) {\ref{grouplegend}}; 
\end{tikzpicture}}
	\caption{Fidelity (accuracy) of neural network and logistic
          regression for varying number of affected features (higher
          values are better).}
	\label{fig:nn_toy_data}
\end{figure*}

\fig{fig:fidelity-multiple-steps} shows the comparison between
sequential and one-shot updates regarding the gradient residual norm
and accuracy on test data. In terms of accuracy, the sequential
updates give only slight performance increases for both methods. For
the gradient residual, however, we observe strong decreases when
applying the updates sequentially, especially for the \diabetis and
\adult dataset. This confirms the results of \cref{thm:thm1}
empirically and presents a special working mode of our approach. As
pointed out in \sect{sect:multiple-steps}, this increase in privacy
budget comes with an increase in runtime for the second-order update:
Performing \num{10} updates instead of one increases the runtime by a
factor \num{10} since the Hessian has to be re-calculated at each
intermediate step.

\subsection{Neural Networks in First Unlearning Scenario}
\label{sec:feat-unle-neur}

Our first unlearning scenario in \sect{sec:unle-sens-names} focuses on
a logistic regression model. This learning model was chosen, because
it provides a strongly convex loss function and thus allows the
application of certified unlearning. However, linear models are
inherently limited in their capabilities, so we also employ a neural
network to the four datasets considered in \sect{sec:unle-sens-names}
for comparison. Note that neural networks generally do not have a
convex loss and therefore do not provide theoretical guarantees for
unlearning in our framework.

In particular, we train a fully connected neural network with two
hidden layers consisting of \num{100} neurons for each of the
datasets. We then remove and replace, respectively, sensitive features
using unlearning as described in \sect{sec:unle-sens-names}. As the
loss of the network is not convex, we cannot use the gradient residual
norm to determine its efficacy or calibrate the noise for certified
unlearning, as done for the logistic regression. Consequently, we drop
the noise term from the loss in this experiment for the neural network
and the logistic regression. Still, we can investigate the accuracy
after unlearning to get a rough reference for the general performance
of our approach on neural networks in this scenario.

\fig{fig:nn_toy_data} shows the accuracy of the neural network and the
logistic regression on the four dataset when unlearning features. Both
models perform very well in this scenario and enable to unlearn 120
features without significant changes of the accuracy. For both models,
the accuracy drops by less than 1\% point on all datasets when
corrections are performed using our first-order and second-order
update. This strong performance demonstrates the capability of our
approach for unlearning with high fidelity. As we shown in
\sect{sec:unle-sens-names}, however, when theoretical guarantess are
enforced, the decrease in accuracy is more pronounced as noise needs
to be added to the model parameters to achieve certification.

We also find that the neural network has a higher accuracy compared to
the logistic regression on all datasets. The non-linear network is
capable of better modeling the underlying data and thus attains a more
accurate prediction. However, the difference to the logistic
regression is marginal and remains below 5\%. For three of the four
datasets (\enron, \drebin, and \adult), it is even less than 1\%. This
result provides an important insight for the unlearning scenario: The
logistic regression and the neural network achieve a comparable
accuracy on the four datasets. Hence, if privacy guarantees are
necessary and a small degradation in performance can be tolerated, the
logistic regression model is actually preferable to the neural
network, despite its inherent limitations.

\subsection{Poisoning Experiment Details}
\label{appdx:poisoning}

In addition to the description in Section~\ref{sec:unle-pois}, we
provide more details about the dataset, CNN architecture and used
unlearning parameters in this section. 

The CIFAR10 dataset~\citep{Kri09} contains \num{50000} images with $32
\times 32 \times 3$ pixels and \num{10} classes representing real-world
objects like vehicles and animals. As our reference model, we train a
convolutional neural network with \num{1.8} million parameters. It
comprises three VGG blocks and two dense layers. 
The network uses \num{128} convolutional filters of size $3 \times 3$,
a pooling size of $2 \times 2$ and the ReLU activation function. We
train it using the Adam optimizer~\citep{KinBa14} with a learning rate
of \num{1e-4}.


\balance

For our approach, we use an unlearning rate~$\tau$ of \num{2e-5} for
the first-order update with batches of size \num{512}. For the
approximation of the inverse Hessian, we use a HVP batch size of
\num{1024}, damping of \num{1e-4} and scale of \num{2e5}
\citep[see][]{AgaBulHaz17}. The batch size defines the number of
unlearning requests that are simultaneously performed and the HVP
batch size denotes the number of training samples that are used in the
computation of the inverse Hessian. \citet{AgaBulHaz17} originally
suggest to repeat the algorithm multiple times and average the results
for a more stable solution but we find that a single repetition is
sufficient in our experiments. Additionally, we introduce a patience
parameter of \num{20} for the norm of the Hessian vector product which
allows for an early stopping when the update norm does not decrease
for a certain amount of iterations.

\end{document}